\documentclass[final,3p,times]{elsarticle}

\usepackage{hyperref}

\usepackage{enumerate}
\usepackage{color}
\usepackage[table,x11names]{xcolor}
\usepackage[linesnumbered,ruled,vlined]{algorithm2e}

\SetCommentSty{mycommfont}
\usepackage{multirow}
\usepackage{stfloats}
\usepackage{url}
\usepackage{graphicx}
\usepackage{amssymb}
\usepackage{subfigure}

\newtheorem{definition}{Definition}
\newtheorem{problem}{Problem}

\gdef\eg{\textit{e.g.}}
\gdef\ie{\textit{i.e.}}

\definecolor{aqua}{rgb}{0.0, 1.0, 1.0}
\graphicspath{{fig/}{expt/}}

\begin{document}

\begin{frontmatter}

\title{Predicting Citywide Crowd Flows Using \\Deep Spatio-Temporal Residual Networks\tnoteref{mytitlenote}}
\tnotetext[mytitlenote]{This paper is an expanded version of \cite{zhang2017aaai}, which has been accepted for presentation at the $31^{st}$ AAAI Conference on Artificial Intelligence (AAAI-17).}

\author[MSR]{Junbo Zhang}
\ead{junbo.zhang@microsoft.com}
\author[MSR,SWJTU,XDU,CAS]{Yu Zheng\corref{mycorrespondingauthor}}
\ead{yuzheng@microsoft.com}
\author[SWJTU]{Dekang Qi\fnref{myfootnote}}
\ead{dekangqi@outlook.com}
\author[XDU]{Ruiyuan Li\fnref{myfootnote}}
\ead{v-ruiyli@microsoft.com}
\author[SWJTU]{Xiuwen Yi\fnref{myfootnote}}
\ead{v-xiuyi@microsoft.com}
\author[SWJTU]{Tianrui Li}
\ead{trli@swjtu.edu.cn}
\address[MSR]{Microsoft Research, Beijing, China}
\address[SWJTU]{School of Information Science and Technology, Southwest Jiaotong University, Chengdu 610031, China}
\address[XDU]{School of Computer Science and Technology, Xidian University, China}
\address[CAS]{Shenzhen Institutes of Advanced Technology, Chinese Academy of Sciences}
\fntext[myfootnote]{The research was done when the third, fourth and fifth authors were interns at Microsoft Research.}
\cortext[mycorrespondingauthor]{Corresponding author}

\begin{abstract}
Forecasting the flow of crowds is of great importance to traffic management and public safety, and very challenging as it is affected by many complex factors, including spatial dependencies (nearby and distant), temporal dependencies (closeness, period, trend), and external conditions (\eg{} weather and events). We propose a deep-learning-based approach, called ST-ResNet, to \textit{collectively} forecast two types of crowd flows (\ie{} inflow and outflow) in each and every region of a city. We design an end-to-end structure of ST-ResNet based on unique properties of spatio-temporal data. More specifically, we employ the residual neural network framework to model the temporal closeness, period, and trend properties of crowd traffic. For each property, we design a branch of residual convolutional units, each of which models the spatial properties of crowd traffic. ST-ResNet learns to dynamically aggregate the output of the three residual neural networks based on data, assigning different weights to different branches and regions. The aggregation is further combined with external factors, such as weather and day of the week, to predict the final traffic of crowds in each and every region. We have developed a real-time system based on \textit{Microsoft Azure Cloud}, called UrbanFlow, providing the crowd flow monitoring and forecasting in Guiyang City of China. In addition, we present an extensive experimental evaluation using two types of crowd flows in Beijing and New York City (NYC), where ST-ResNet outperforms nine well-known baselines. 
\end{abstract}


\end{frontmatter}


%
\section{Introduction}
Predicting crowd flows in a city is of great importance to traffic management, risk assessment, and public safety \cite{Zheng2014AToISaTT}. For instance, massive crowds of  people streamed into a strip region at the 2015 New Year's Eve celebrations in Shanghai, resulting in a catastrophic stampede that killed  36 people. In mid-July of 2016, hundreds of ``Pokemon Go'' players ran through New York City's Central Park in hopes of catching a particularly rare digital monster, leading to a dangerous stampede there. If one can predict the crowd flow in a region, such tragedies can be mitigated or prevented by utilizing emergency mechanisms, such as conducting traffic control, sending out warnings, or evacuating people, in advance. 

\begin{figure}[!htbp]%
\centering
\subfigure[{Inflow and outflow}]{\label{fig:in_out_flows}\includegraphics[width=.25\linewidth]{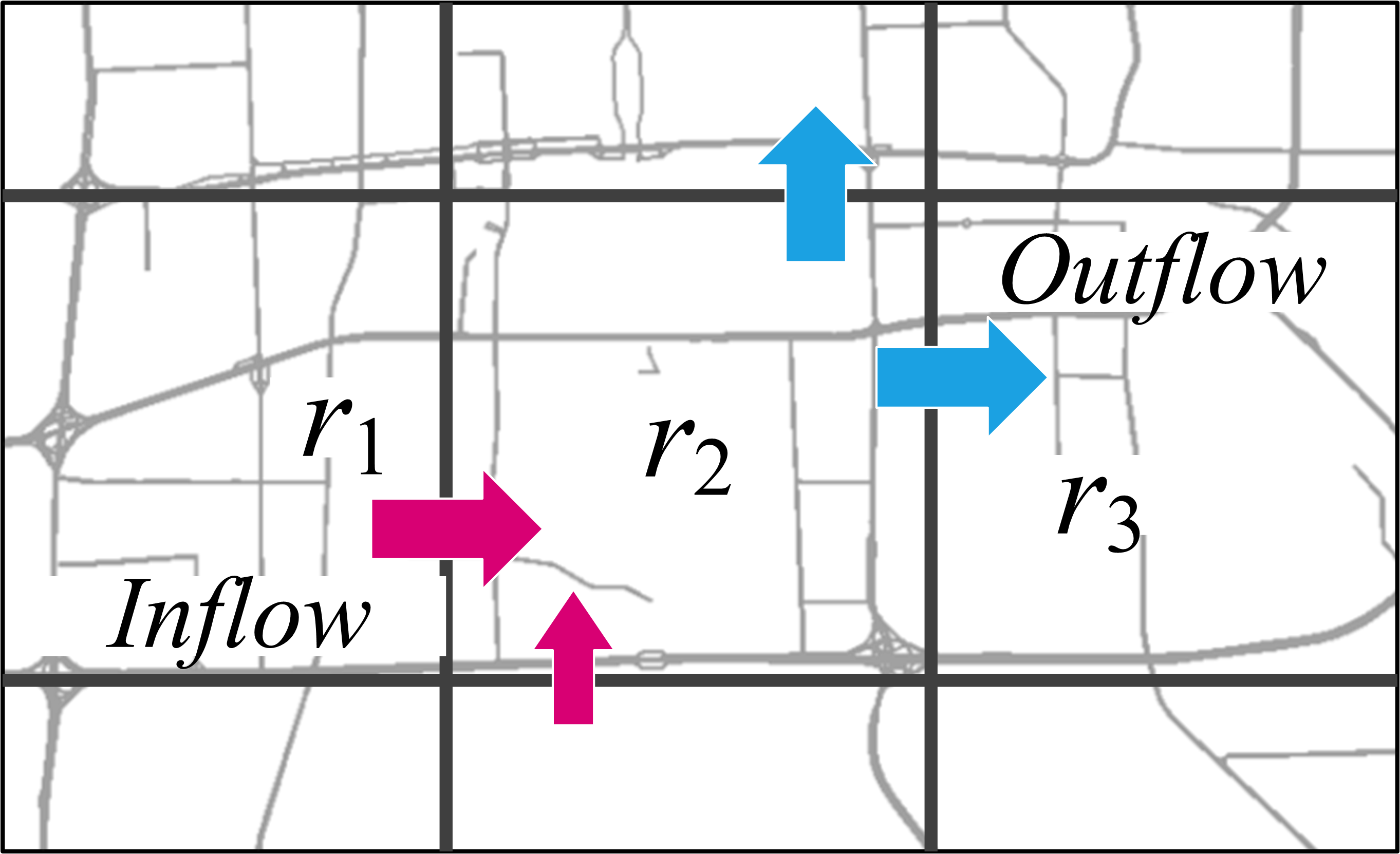}}
\hspace{2em}
\subfigure [Measurement of flows]{\label{fig:calc}\includegraphics[width=.25\linewidth]{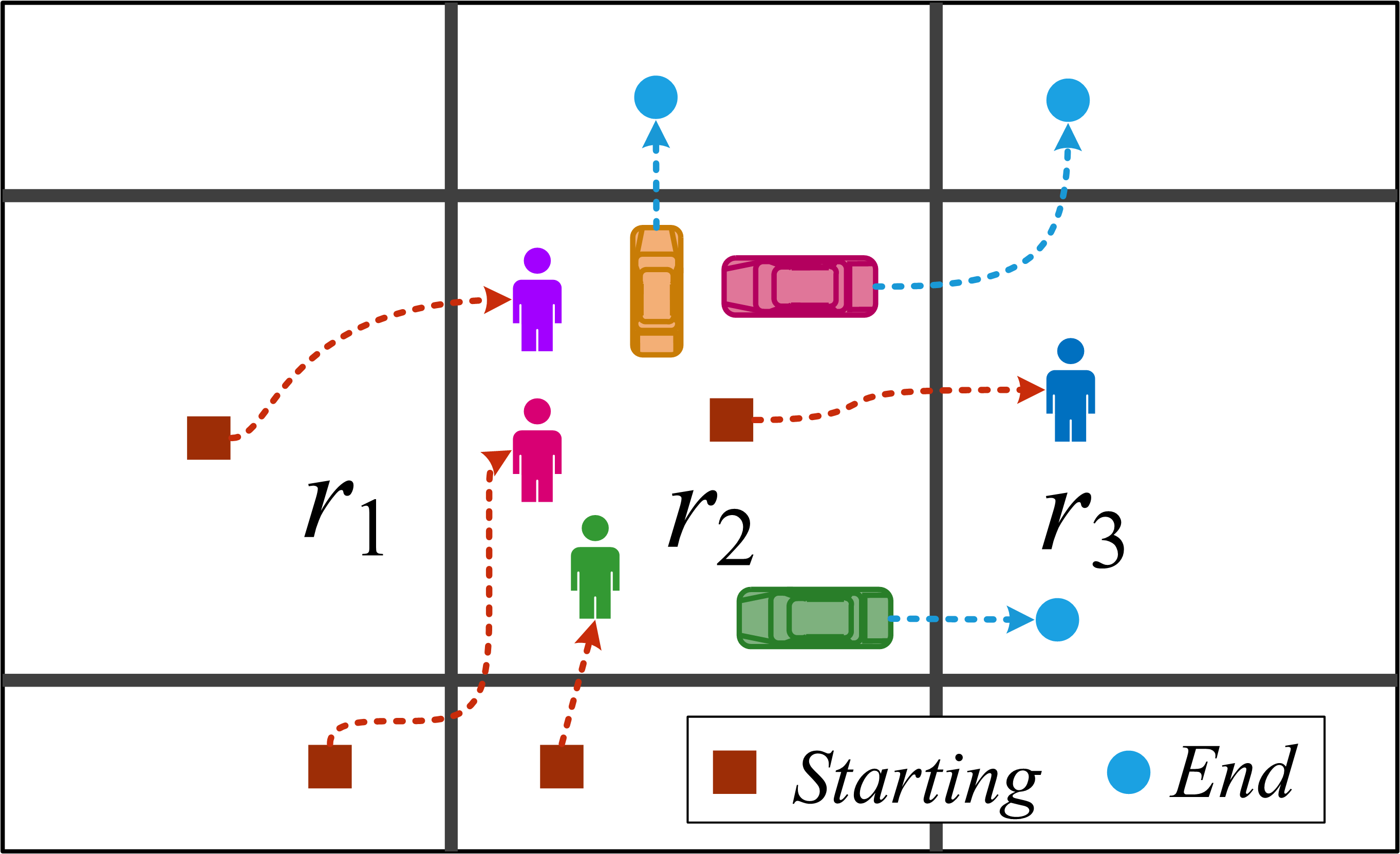}}
\caption{Crowd flows in a region}
\end{figure}

In this paper, we predict two types of crowd flows \cite{Zhang2016}: inflow and outflow, as shown in Figure~\ref{fig:in_out_flows}. 
Inflow is the total traffic of crowds entering a region from other places during a given time interval. 
Outflow denotes the total traffic of crowds leaving a region for other places during a given time interval. Both types of flows track the transition of crowds between regions. Knowing them is very beneficial for risk assessment and traffic management. 
Inflow/outflow can be measured by the number of pedestrians, the number of cars driven nearby roads, the number of people traveling on public transportation systems (\eg{}, metro, bus), or \textit{all of them together} if data is available. 
Figure~\ref{fig:calc} presents an illustration. We can use mobile phone signals to measure the number of pedestrians, showing that the inflow and outflow of $r_2$ are $(3, 1)$, respectively. Similarly, using the GPS trajectories of vehicles, two types of flows are $(0,3)$, respectively. Therefore, the total inflow and outflow of $r_2$ are $(3, 4)$, respectively. Apparently, predicting crowd flows can be viewed as a kind of spatio-temporal prediction problem \cite{Zheng2014AToISaTT}. 

Deep learning \cite{lecun2015deep} has been used successfully in many applications, and is considered to be one of the most cutting-edge artificial intelligence (AI) techniques. 
Exploring these techniques for spatio-temporal data is of great importance to a series of various spatio-temporal applications, including urban planning, transportation, the environment, energy, social, economy, public safety and security \cite{Zheng2014AToISaTT}. 
Although two main types of deep neural networks (DNNs) have considered a sort of spatial or temporal property: 1) convolutional neural networks (CNNs) for capturing spatial structures; 2) recurrent neural networks (RNNs) for learning temporal dependencies. 
It is still very challenging to apply these existing AI techniques for such spatio-temporal prediction problem because of the following three complex factors: 

\begin{enumerate}
\item Spatial dependencies.
\vspace{-10pt}
\begin{description}
\item[Nearby] The inflow of Region $r_2$ (shown in Figure~\ref{fig:in_out_flows}) is affected by outflows of \textit{nearby} regions (like $r_1$). Likewise, the outflow of $r_2$ would affect inflows of other regions (\eg{}, $r_3$). The inflow of region $r_2$ would affect its own outflow as well. 
\vspace{-6pt}
\item[Distant] The flows can be affected by that of \textit{distant} regions. For instance, people who lives far away from the office area always go to work by metro or highway, implying that the outflows of \textit{distant} regions directly affect the inflow of the office area. 
\end{description}
\item Temporal dependencies. 
\vspace{-10pt}
\begin{description}
\item[Closeness] The flow of crowds in a region is affected by recent time intervals, both near and far. For instance, a traffic congestion occurring at 8:00am will affect that of 9:00am. And the crowd flows of today's $16^{th}$ time interval\footnote{Assume that half-hour is a time interval, $16^{th}$ time interval means 7:30am - 8:00am.} is more similar to that of yesterday's $16^{th}$ time interval than that of today's $20^{th}$ time interval. 
\vspace{-6pt}
\item[Period] Traffic conditions during morning rush hours may be similar on consecutive weekdays, repeating every 24 hours. 
\vspace{-6pt}
\item[Trend] Morning rush hours may gradually happen later as winter comes. When the temperature gradually drops and the sun rises later in the day, people get up later and later. 
\end{description}
\item External influence. Some external factors, such as weather conditions and events may change the flow of crowds tremendously in different regions of a city. For example, a thunderstorm affects the traffic speed on roads and further changes the flows of regions. 
\end{enumerate}

To tackle above mentioned challenges, we here explore DNNs for spatio-temporal data, and propose a deep spatio-temporal residual network (ST-ResNet) to \textit{collectively} predict inflow and outflow of crowds in every region. 
Our contributions are five-fold:
\begin{itemize}
\item ST-ResNet employs convolution-based residual networks to model both \textit{nearby} and \textit{distant} spatial dependencies between any two regions in a city, while ensuring the model's prediction accuracy is not comprised by the deep structure of the neural network. 
\vspace{-6pt}
\item We summarize the temporal properties of crowd flows into three categories, consisting of temporal \textit{closeness}, \textit{period}, and 
\textit{trend}. ST-ResNet uses three different residual networks to model these properties, respectively. 
\vspace{-6pt}
\item ST-ResNet dynamically aggregates the output of the three aforementioned networks, assigning different weights to different branches and regions. The aggregation is further combined with external factors (\eg{}, weather). 
\vspace{-6pt}
\item We evaluate our approach using Beijing taxicabs' trajectories and meteorological data, and NYC bike trajectory data. The results demonstrate the advantages of our approach compared with 9 baselines. 
\vspace{-6pt}
\item We develop a real-time crowd flow monitoring \& forecasting system using ST-ResNet. And our solution is based on the \textit{Cloud} and GPU servers, providing powerful and flexible computational environments.  %
\end{itemize}

The rest of this paper is organized as follows. In Section~\ref{sec:preliminary}, we formally describe the crowd flow prediction problem. 
Section~\ref{sec:system_arch} overviews the architecture of our proposed system. 
Section~\ref{sec:method} describes the DNN-based prediction model used. 
We present the evaluation in Section~\ref{sec:expt} and summarized the related work in Section~\ref{sec:related_work}. 

The differences between this paper and our earlier work \cite{zhang2017aaai} are four aspects.  
First, we have deployed a cloud-based system that continuously forecasts the flow of taxicabs in each and every region of Guiyang City in China, showcasing the capability of ST-ResNet in handling real-world problems. The implementation of the cloud-based system is also introduced in Section~\ref{sec:system_arch} of this paper. 
Second, we extend ST-ResNet from a one-step ahead prediction to a multi-step ahead prediction, enabling the prediction of crowd flows over a farther future time (Section~\ref{sec:alg}). 
Third, we conduct more comprehensive experiments on crowd flow prediction, showcasing the effectiveness and robustness of ST-ResNet: i) comparing our method with more advanced baselines (e.g., three different variants of recurrent neural networks) (Section~\ref{sec:expt:single}); ii) testing more network architectures for ST-ResNet (Section~\ref{sec:expt:param}); iii) adding the experiments of multi-step ahead prediction (Section~\ref{sec:expt:multi}); iv) discussing the performance of our method changing over different cloud resources (Section~\ref{sec:expt:cloud}). 
Fourth, we have explored more related works (in Section~\ref{sec:related_work}), clarifying the differences and connections to the-state-of-the-art. This helps better position our research in the community.

\section{Preliminary}\label{sec:preliminary}
We first briefly revisit the crowd flow prediction problem \cite{Zhang2016} and then introduce deep residual learning \cite{He2016apa}. 
\subsection{Formulation of Crowd Flow Prediction Problem}
\begin{figure}[!htbp]%
\centering
\subfigure[{Map segmentation}]{\label{fig:gridmap}\includegraphics[height=.16\linewidth]{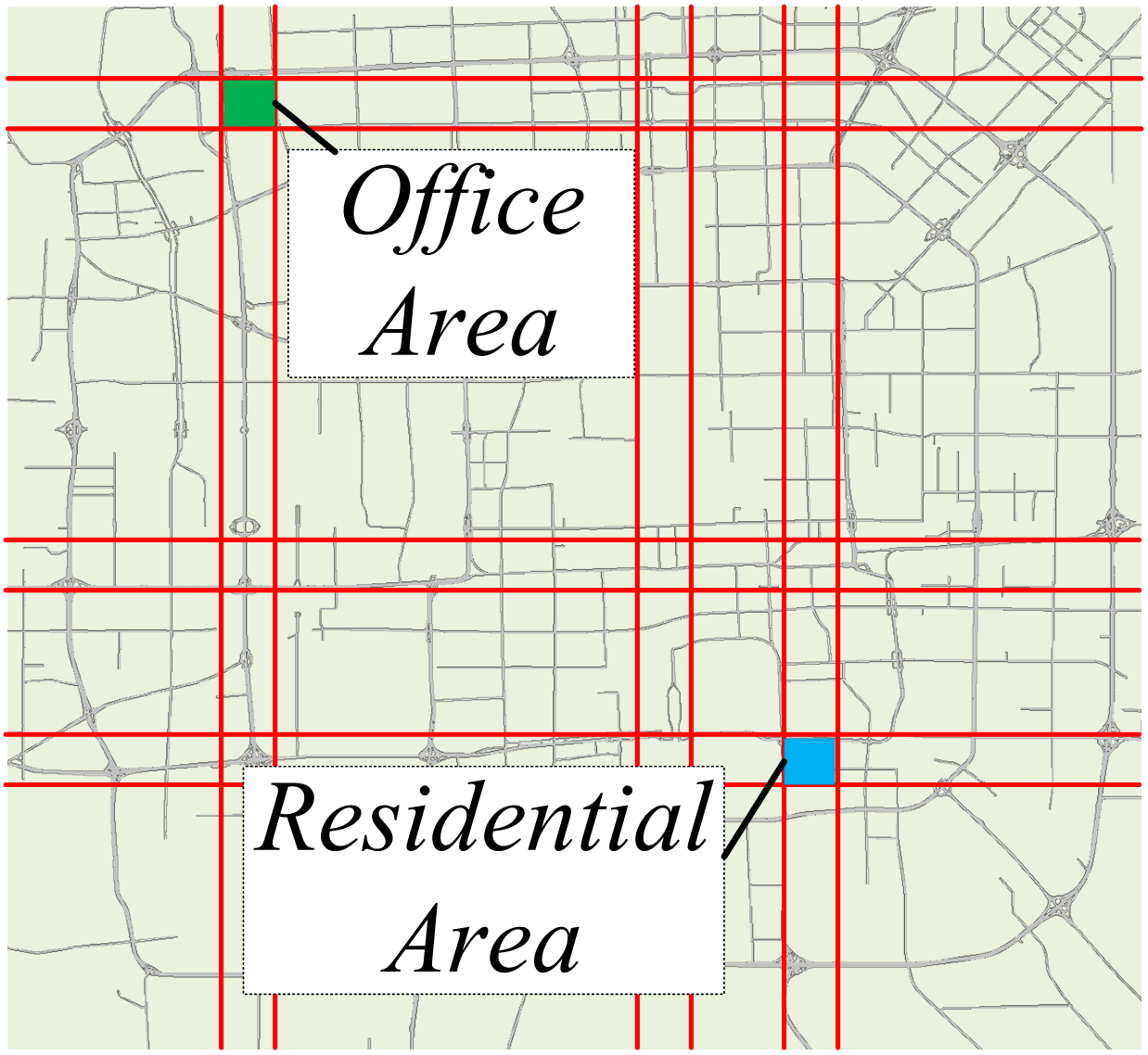}}
\hspace{2em}
\subfigure [Inflow matrix]{\label{fig:heatmap}\includegraphics[height=.16\linewidth]{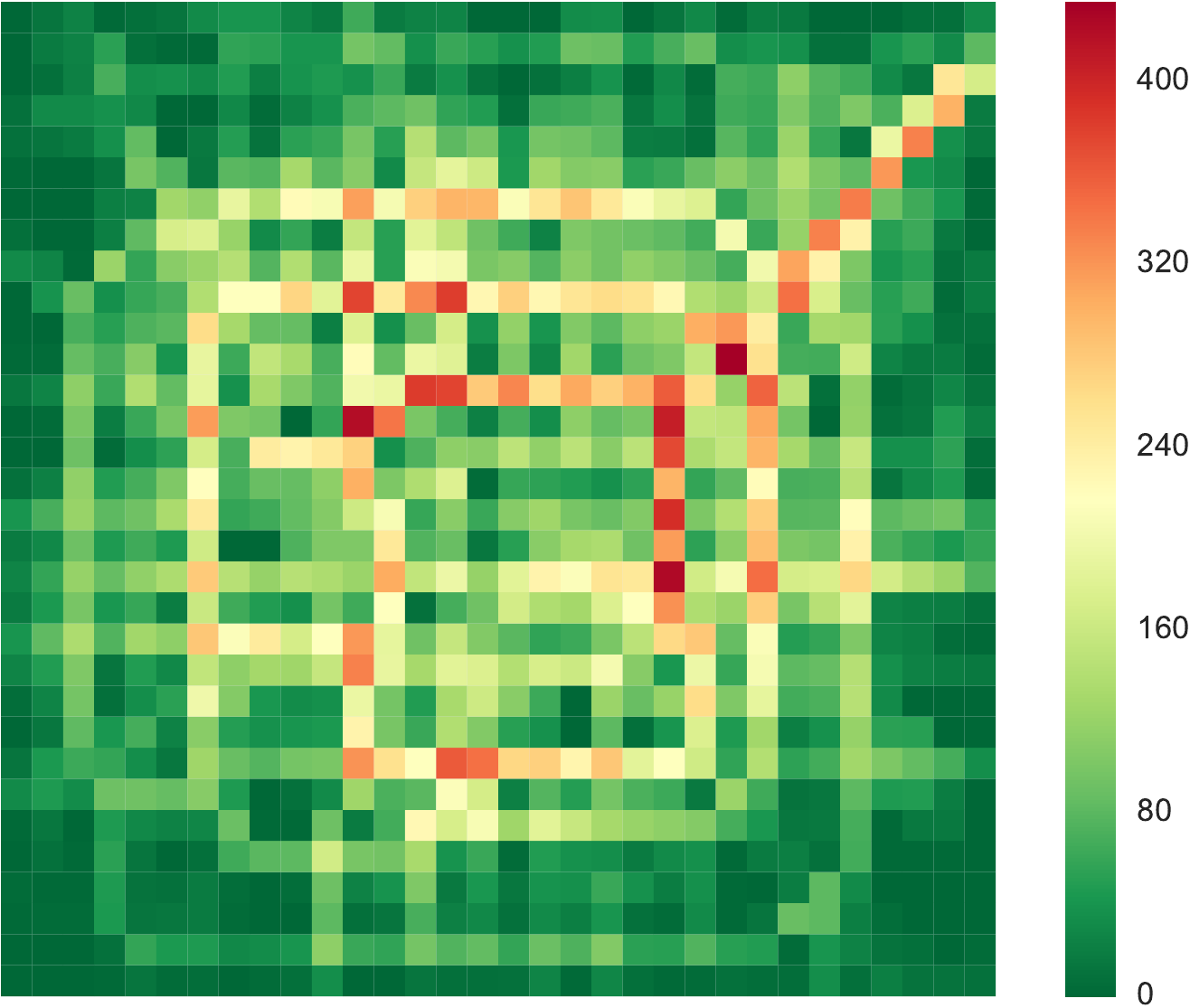}}
\vspace{-10pt}
\caption{Regions in Beijing: (a) Grid-based map segmentation; (b) inflows in every region of Beijing}
\end{figure}

\begin{definition}[Region \cite{Zhang2016}]\label{def:region}
There are many definitions of a location in terms of different granularities and semantic meanings. In this study, we partition a city into an $I \times J$ grid map based on the longitude and latitude where a grid denotes a region, as shown in Figure~\ref{fig:gridmap}.
\end{definition}

\vspace{-20pt}

\begin{definition}[Inflow/outflow \cite{Zhang2016}]\label{def:flow}
Let $\mathbb P$ be a collection of trajectories at the $t^{th}$ time interval. 
For a grid $(i, j)$ that lies at the $i^{th}$ row and the $j^{th}$ column, the inflow and outflow of the crowds at the time interval $t$ are defined respectively as 
\begin{eqnarray}
x_t^{in, i, j} &=& \sum\limits_{Tr \in \mathbb P} |\{k>1 | g_{k-1} \not\in (i, j) \wedge g_k \in (i, j)\}| \nonumber \\
x_t^{out, i, j} &=& \sum\limits_{Tr \in \mathbb P} |\{k\geq 1 | g_k \in (i, j) \wedge g_{k+1} \not\in (i, j)\}| \nonumber
\end{eqnarray}
where $Tr: g_1 \rightarrow g_2 \rightarrow \cdots  \rightarrow g_{|Tr|}$ is a trajectory in $\mathbb P$, and
$g_k$ is the geospatial coordinate; $g_k \in (i, j)$ means the point $g_k$ lies within grid $(i, j)$, and vice versa; $|\cdot|$ denotes the cardinality of a set. 
\end{definition}
At the $t^{th}$ time interval, inflow and outflow in all $I \times J$ regions can be denoted as a tensor 
$\mathbf X_t \in \mathbb R ^{2\times I\times J}$ where $(\mathbf X_t)_{0,i, j}=x_t^{in, i, j}$, $(\mathbf X_t)_{1,i, j}=x_t^{out, i, j}$. 
The inflow matrix is shown in Figure~\ref{fig:heatmap}. 

Formally, for a dynamical system over a spatial region represented by a $I \times J$ grid map, there are 2 types of flows in each grid over time. 
Thus, the observation at any time can be represented by a tensor $\mathbf X \in \mathbb R ^{2 \times I\times J}$. 
\begin{problem}
Given the historical observations $\{\mathbf X_t| t=0, \cdots, n-1\}$, predict $\mathbf X_n$.
\end{problem}

\subsection{Deep Residual Learning}
Deep residual learning \cite{He2015apa} allows convolution neural networks to have a super deep structure of 100 layers, even over-1000 layers. 
And this method has shown state-of-the-art results on multiple challenging recognition tasks, including image classification, object detection, segmentation and localization \cite{He2015apa}. 

Formally, a residual unit with an identity mapping \cite{He2016apa} is defined as:
\begin{equation}
\mathbf X^{(l+1)} = \mathbf X^{(l)} + \mathcal F(\mathbf X^{(l)})
\end{equation}
where $\mathbf X^{(l)}$ and $\mathbf X^{(l+1)}$ are the input and output of the $l^{th}$ residual unit, respectively; $\mathcal F$ is a residual function, \eg{}, a stack of two $3\times 3$ convolution layers in \cite{He2015apa}. 
The central idea of the residual learning is to learn the additive residual function $\mathcal F$ with respect to $\mathbf X^{(l)}$ \cite{He2016apa}. 

\section{System Architecture}\label{sec:system_arch}
\begin{figure}[!htbp]
\centering
\includegraphics[width=0.6\linewidth]{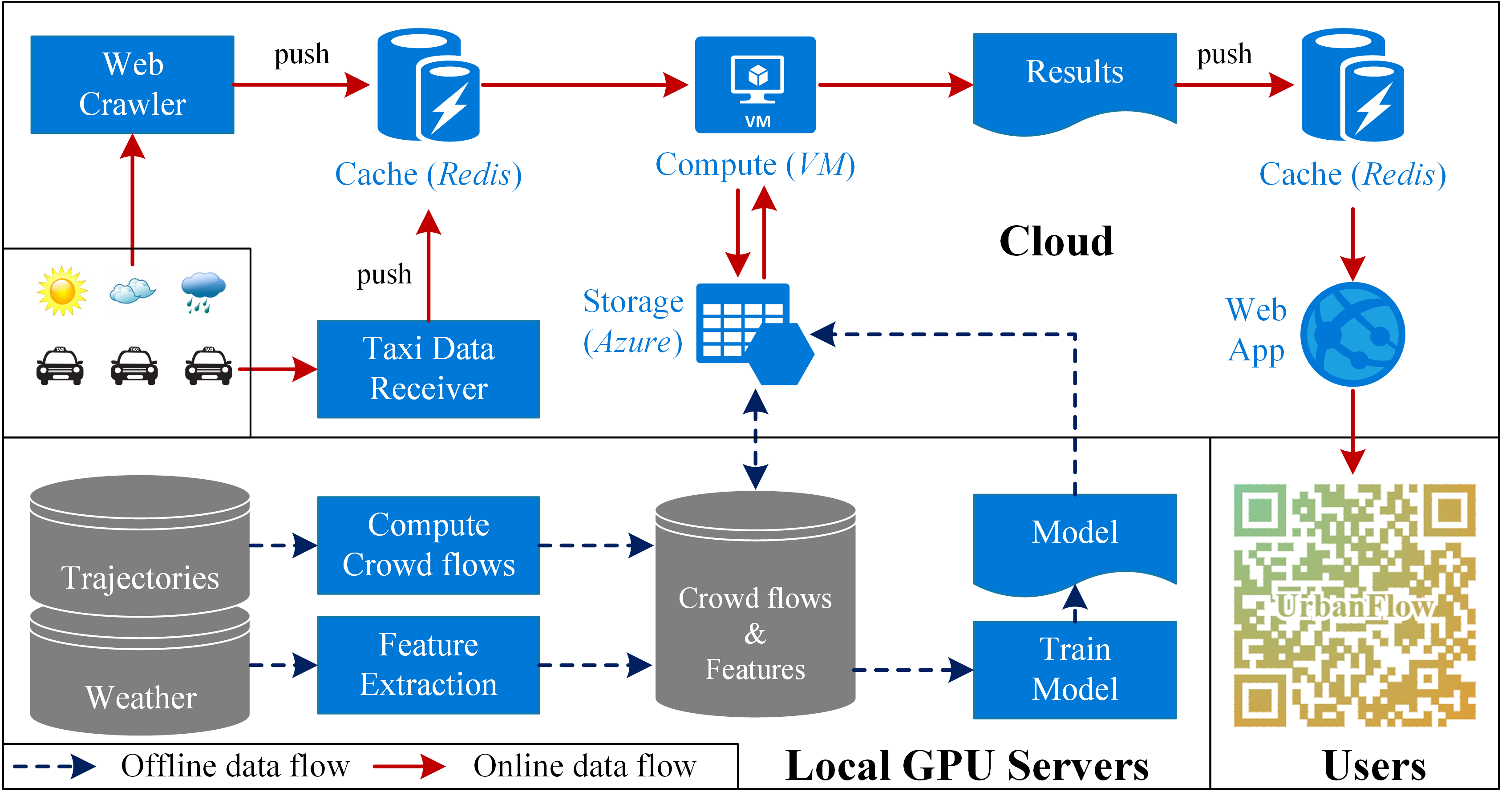}
\caption{System Framework} %
\label{fig:system_framework}
\end{figure}

Figure~\ref{fig:system_framework} presents the framework of our system, which consists of three major parts: \textit{local GPU servers}, and the \textit{Cloud}, and \textit{users} (\eg{}, website and QR Code), resulting in offline and online data flows, respectively. 
The local GPU servers store historical observations, such as taxi trajectories, meteorological data. 
The Cloud receive real-time data, including real-time traffic data (\eg{} trajectories) within a time interval as well as meteorological data. 
The users access the inflow/outflow data, displaying them on websites or smart phone via scanning QR code. 

\subsection{The Cloud}\label{sec:cloud}
The cloud continuously receives GPS trajectories of taxicabs and crawls meteorological data, and then caches them into a \textit{redis}. 
A virtual machine (VM) (or VMs) on the Cloud pulls these data from the \textit{redis}, and then computes crowd flows according to GPS trajectories for each and every region of a city. Meanwhile, the VM extracts the features from meteorological data, event data and others. Afterwards, the VM stores the crowd flow data and extracted features into the \textit{storage} (a part of the VM). 
To save the resource on the cloud (more storages need more expensive payment), we only store the crowd flow data and features in past two days. Historical data can be moved to local servers periodically. 

We use Azure platform as a service (PaaS). Table~\ref{tab:cloud} details the Azure\footnote{https://azure.microsoft.com} resources for our system as well as the price\footnote{https://azure.microsoft.com/en-us/pricing}. 
We employ a VM\footnote{To accelerate the computation, one also can choose a more powerful VM, such as \textit{D4 standard} of Azure that has 8 cores and 28 GB memory with $\$ 0.616$ per hour.}, called \textit{A2 standard in Azure}, that has 2 cores and 3.5 GB memory for forecasting the crowd flows in near future. 
The website and web service share a \textit{App Service}, given the potential heavy accesses by many users. 
As the  historical data is stored in local servers, a 6 GB \textit{Redis Cache} is enough for caching the real-time trajectories in past half-hour, crowd flow data \& extracted features in past two days, and inferred results. 
\begin{table}[!htbp]
\tabcolsep 8pt
\begin{center}
\caption{The Azure resources used for our system}
\label{tab:cloud} 
\begin{tabular}{c|c|c}
\hline
\hline
Azure Service & Configuration & Price \\
\hline
App Service & Standard, 4 cores, 7GB memory & $\$ 0.400/hour$ \\
Virtual Machine & A2 standard, 2 cores, 3.5 GB memory & $\$ 0.120/hour$\\
Redis Cache & P1 premium, 6GB & $\$ 0.555/hour$ \\
\hline
\hline
\end{tabular}
\end{center}
\end{table}
\vspace{-20pt}
\subsection{Local GPU Servers}
Although all the jobs can be done on the cloud, GPU services on the Cloud is not supported in some areas (\eg{}, China). On the other hand, we need to pay for other cloud services, like storages and I/O bandwidths. Saving unnecessary cost is vital for a research prototype. 
In addition, migrating massive data from local servers up to the cloud is time-consuming given the limited network bandwidth. For instance, the historical trajectories can be hundreds of Gigabytes, even Terabytes, leading to a very long time for copying the data from local servers to the cloud. 

Therefore, we employ a hybrid framework that combines local GPU servers with the cloud. 
\textit{Local GPU servers} mainly handle the offline training (learning), including three tasks: 
\begin{itemize}
\item Converting trajectories into inflow/outflow: we first use the massive historical trajectories and employ a calculation module to get crowd flow data, then store them in local. 
\item Extracting features from external data: we fist collect external data (\eg{} weather, holiday events) from different data sources and fit them into a feature extraction module to get continuous or discrete features, and then store them in local. 
\item Training model: we use above generated crowd flows and external features to train a predictive model via our proposed ST-ResNet, and then upload the learned model to the cloud. Note that as the dynamic crowd flows and features are stored in the Storage (Azure), we sync up the online data to local servers before each training processing. In this way, we are agile to try new ideas (\eg{} re-train the model) while greatly reducing expense for a research prototype. 
\end{itemize}

\subsection{User Interface}
\begin{figure}[!htbp]
\centering
\includegraphics[width=.8\linewidth]{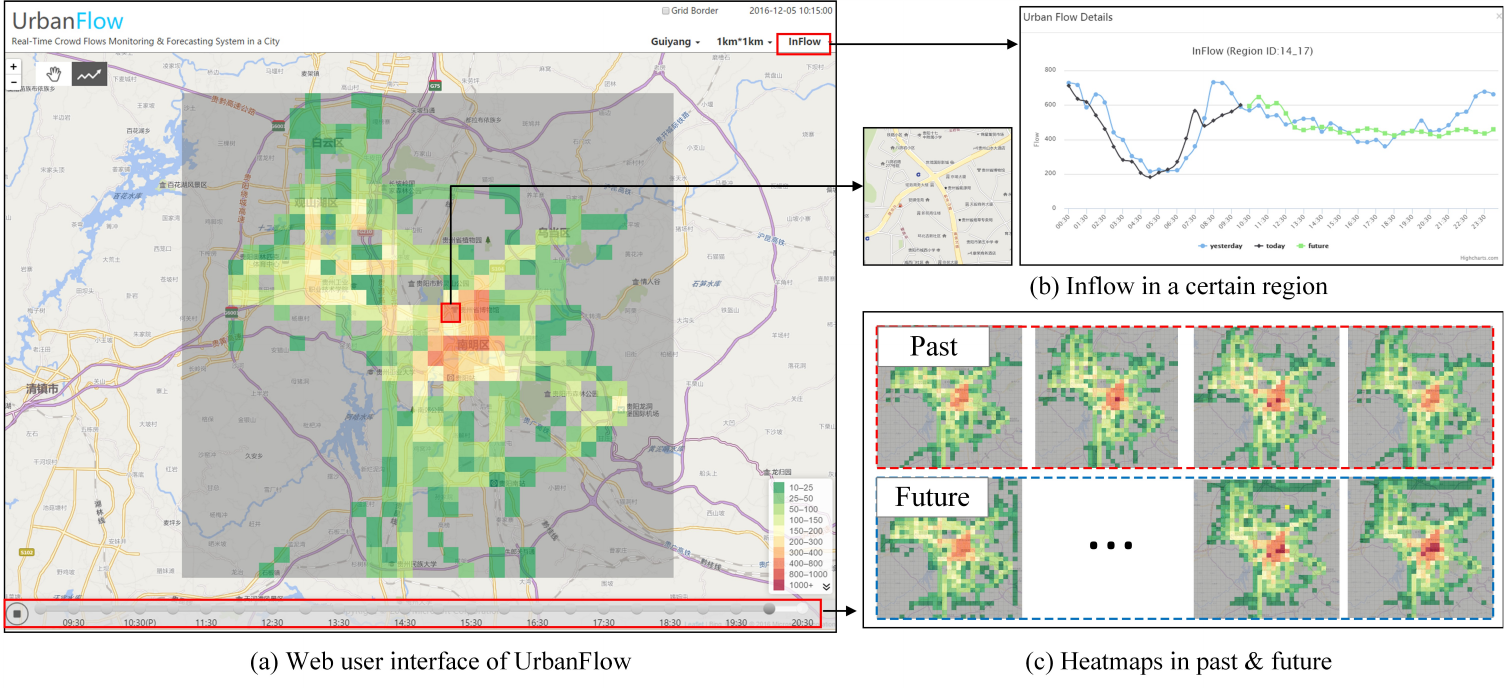}
\vspace{-10pt}
\caption{Web user interface of UrbanFlow}
\label{fig:urbanflow}
\end{figure}

Figure~\ref{fig:urbanflow}(a) presents the website of UrbanFlow \cite{urbanflow}, where each grid on the map stands for a region and the number associated with it denotes its inflow or outflow of crowds. The user can view inflow or outflow via the top-right button named ``InFlow/OutFlow''. The smaller the number is, the crowd flow is sparser. The color of each grid is determined in accordance with its crowd flows, \eg{}, ``red'' means ``dense'' crowd flow and ``green'' means ``sparse'' crowd flow. The top-right corner of the website shows the buttons which can switch between different types of flows. A user can select any grid (representing a region) on the website and click it to see the region's detailed flows, as shown in Figure~\ref{fig:urbanflow}(b) where blue, black, and green curves indicate flows of yesterday, past, and future times at today, respectively.
The bottom of the website shows a few sequential timestamps. The heatmap at a certain timestamp will be shown in the website when a user clicks the associated timestamp. Intuitively, the user can watch the movie-style heatmaps (Figure~\ref{fig:urbanflow}(c)) by clicking ``play button'' at the bottom-left of Figure~\ref{fig:urbanflow}(a). 
At present, we apply UrbanFlow to the area of Guiyang City, China \footnote{http://urbanflow.sigkdd.com.cn/}. 

\section{Deep Spatio-Temporal Residual Networks}\label{sec:method}
\begin{figure}[!htbp]
\centering
\includegraphics[width=0.5\linewidth]{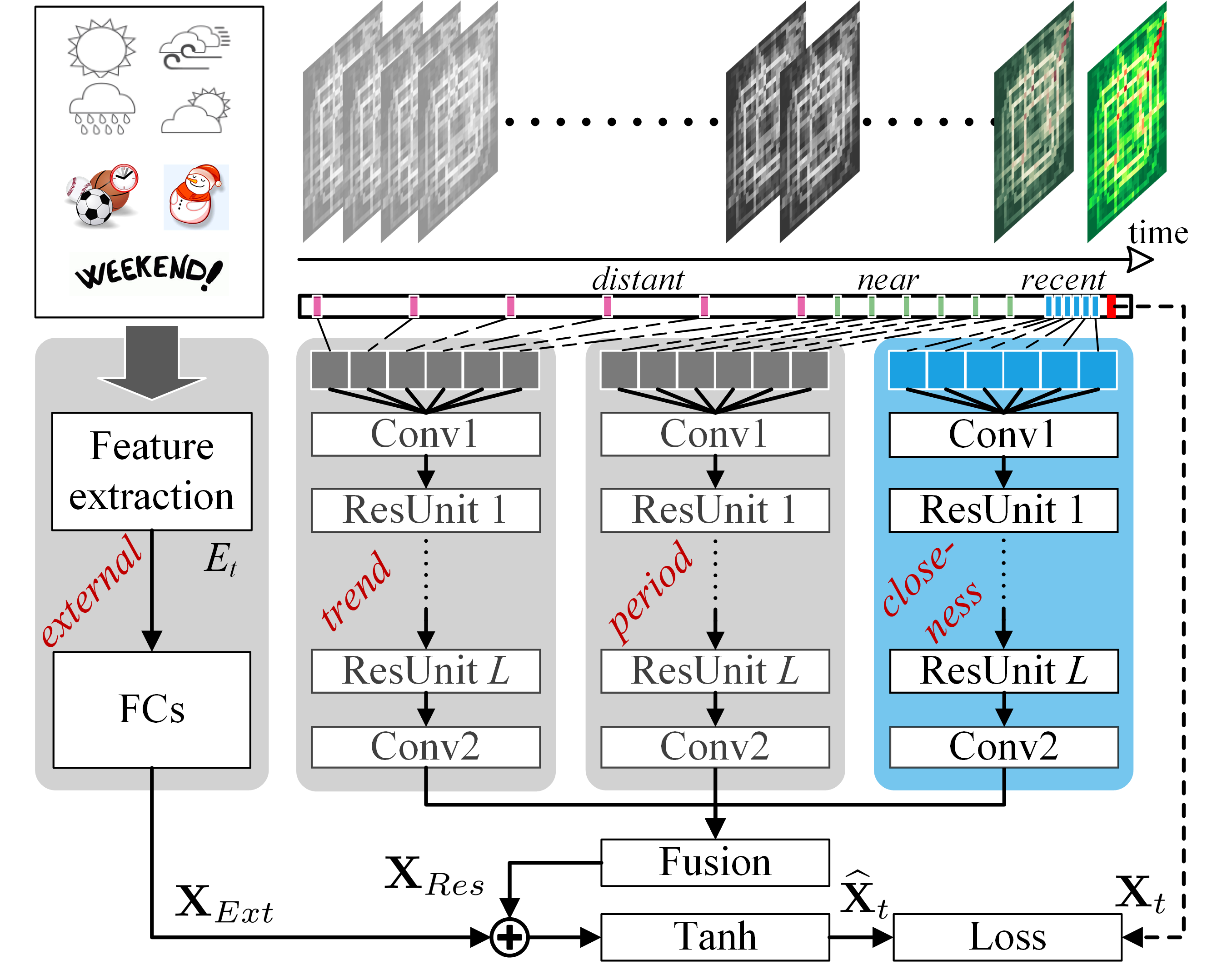}
\caption{ST-ResNet architecture. Conv: Convolution; ResUnit: Residual Unit; FC: Fully-connected.}
\label{fig:STResNet}
\end{figure}

Recurrent neural networks (RNNs), like long-short term memory (LSTM), is capable of learning long-range temporal dependencies. 
Using RNNs, however, to model temporal \textit{period} and \textit{trend}, it needs very long input sequences (\eg{}, $\rm{length}=1344$)\footnote{Assume that half-an-hour is a time interval, 4-week sequence's length is equal to $48 \times 7 \times 4 = 1344.$}, which makes the whole training processing non-trivial (see Section~\ref{sec:expt:single} for empirical evaluation).
According to the ST domain knowledge, we know that only a few previous keyframes influence the next keyframe. Therefore, we leverage temporal \textit{closeness}, \textit{period}, \textit{trend} to select keyframes for modeling. 
Figure~\ref{fig:STResNet} presents the architecture of ST-ResNet, which is comprised of four major components modeling temporal \textit{closeness}, \textit{period}, \textit{trend}, and \textit{external} influence, respectively. 

As illustrated in the top-right part of Figure~\ref{fig:STResNet}, we first turn inflow and outflow throughout a city at each time interval into a 2-channel image-like matrix respectively, using the approach introduced in Definitions 1 and 2. 
We then divide the time axis into three fragments, denoting \textit{recent} time, \textit{near} history and \textit{distant} history. The 2-channel flow matrices of intervals in each time fragment are then fed into the first three components separately to model the aforementioned three temporal properties: \textit{closeness}, \textit{period} and \textit{trend}, respectively. The first three components share the same network structure with a convolutional neural network followed by a Residual Unit sequence. Such structure captures the spatial dependency between nearby and distant regions. In the \textit{external} component, we manually extract some features from external datasets, such as weather conditions and events, feeding them into a two-layer fully-connected neural network. The outputs of the first three components are fused as $\mathbf X_{Res}$ based on parameter matrices, which assign different weights to the results of different components in different regions. $\mathbf X_{Res}$ is further integrated with the output of the external component $\mathbf X_{Ext}$. Finally, the aggregation is mapped into $[-1, 1]$ by a Tanh function, which yields a faster convergence than the standard logistic function in the process of back-propagation learning \cite{lecun2012efficient}.

\subsection{Structures of the First Three Components}
\begin{figure}[!htbp]
\centering
\subfigure[Convolutions  ]{\label{fig:holiday}\includegraphics[width=.35\linewidth]{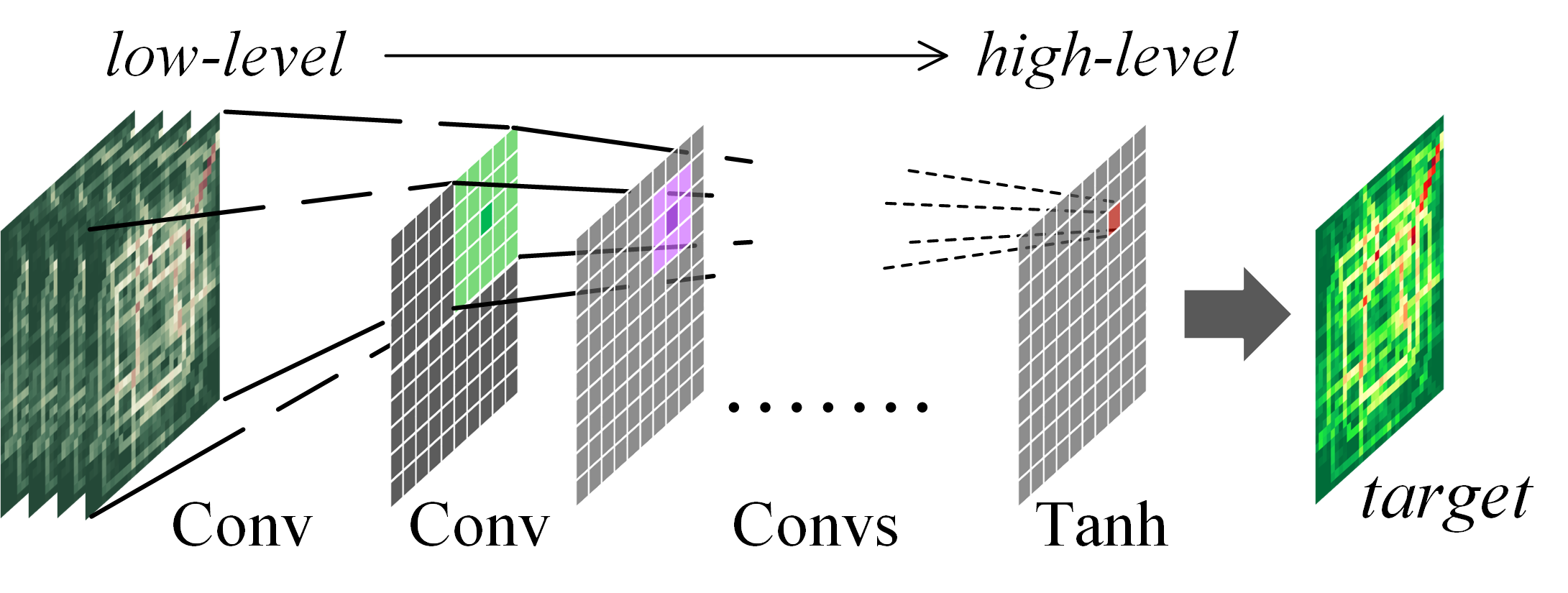}}
\subfigure[Residual Unit]
{\label{fig:weather}\includegraphics[width=.45\linewidth]{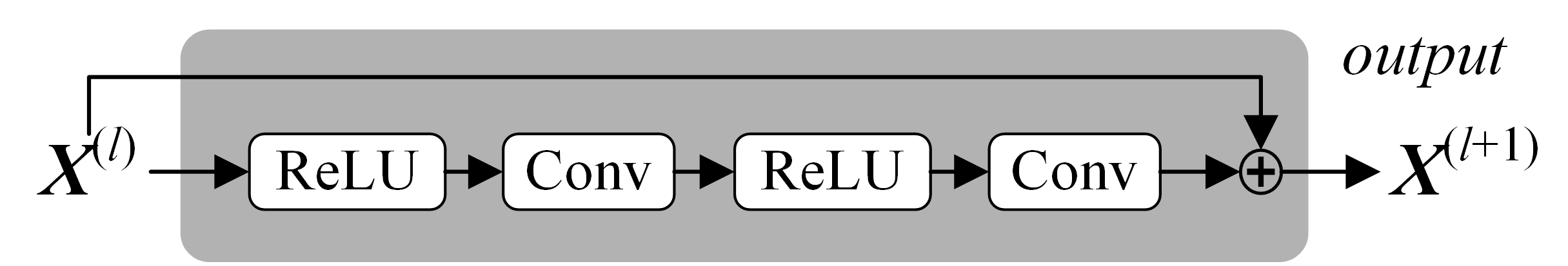}}
\caption{Convolution and residual unit}
\label{fig:ResUnit}
\end{figure}

The first three components (\ie{} \textit{closeness}, \textit{period}, \textit{trend}) share the same network structure, which is composed of two sub-components: convolution and residual unit, as shown in Figure~\ref{fig:ResUnit}.

\noindent\textit{\textbf{Convolution.}} 
A city usually has a very large size, containing many regions with different distances. 
Intuitively, the flow of crowds in nearby regions may affect each other, which can be effectively handled by the convolutional neural network (CNN) that has shown its powerful ability to hierarchically capture the spatial structural information \cite{LeCun1998PotI}. 
In addition, subway systems and highways connect two locations with a far distance, leading to the dependency between distant regions. In order to capture the spatial dependency of any region, we need to design a CNN with many layers because one convolution only accounts for spatial near dependencies, limited by the size of their kernels. 
The same problem also has been found in the video sequence generating task where the input and output have the same resolution \cite{Mathieu2015apa}. 
Several methods have been introduced to avoid the loss of resolution brought about by subsampling while preserving distant dependencies \cite{Long2015}. Being different from the classical CNN, we do not use subsampling, but only convolutions \cite{Jain2007}. As shown in Figure~\ref{fig:ResUnit}(a), there are three multiple levels of feature maps that are connected with a few convolutions. 
We find that a node in the high-level feature map depends on nine nodes of the middle-level feature map, those of which depend on all nodes in the lower-level feature map (\ie{} input). It means one convolution naturally captures spatial near dependencies, and a stack of convolutions can further capture distant even citywide dependencies. 

The \textit{closeness} component of Figure~\ref{fig:STResNet} adopts a few 2-channel flows matrices of intervals in the recent time to model temporal \textit{closeness} dependency. 
Let the recent fragment be $[\mathbf X_{t-{l_c}}, \mathbf X_{t-{(l_c -1)}} ,\cdots, \mathbf X_{t-1}]$, which is also known as the \textit{closeness} dependent sequence. 
We first concatenate them along with the first axis (\ie{} time interval) as one tensor $\mathbf X_c^{(0)} \in \mathbb R^{2l_c \times I \times J}$, which is followed by a convolution (\ie{} \texttt{Conv1} shown in Figure~\ref{fig:STResNet}) as:
\begin{equation}\label{eq:conv}
\mathbf X_{c}^{(1)} = f \left(W^{(1)}_{c} *\mathbf X_c^{(0)}+ b^{(1)}_{c} \right) \nonumber
\end{equation}
where $*$ denotes the convolution in a convolutional operator; $f$ is an activation function, \textit{e.g.} the rectifier $f(z) := \max (0,z)$ \cite{Krizhevsky2012}; $W^{(1)}_{c}, b_c^{(1)}$ are the learnable parameters in the first layer.  

\begin{figure}[!htbp]
\centering
\includegraphics[width=0.4\linewidth]{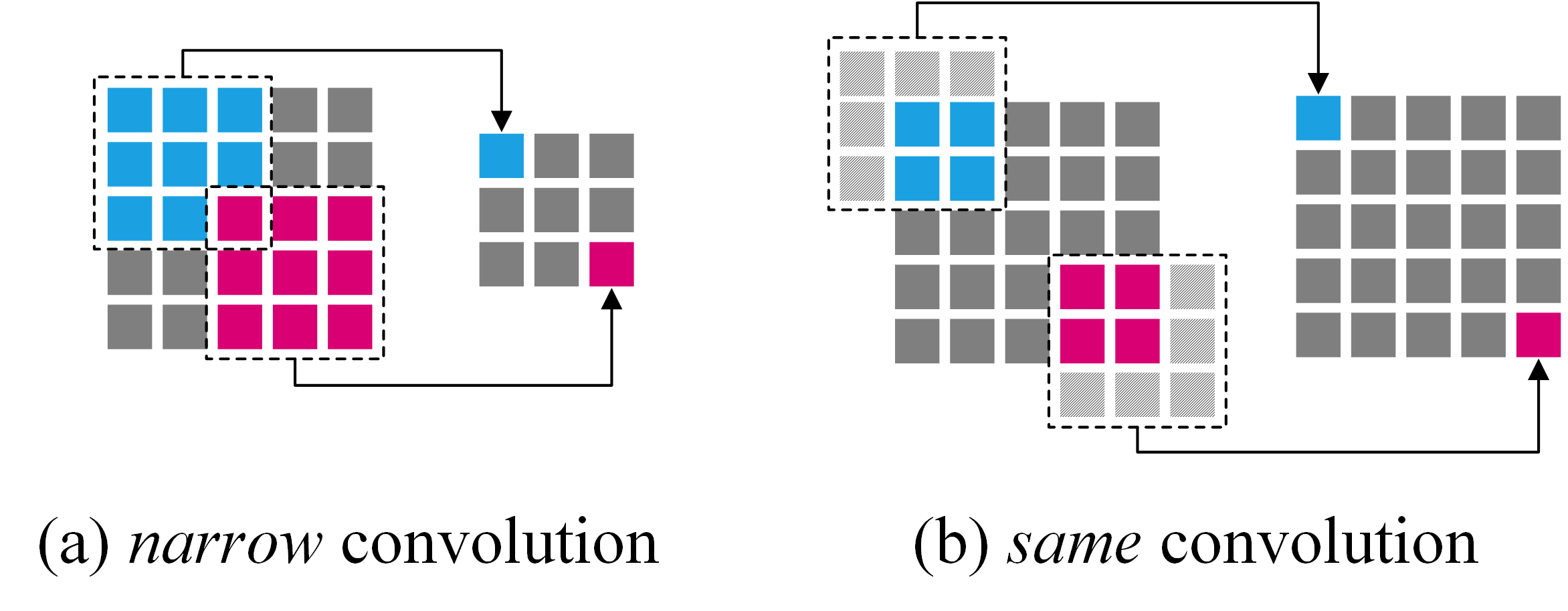}
\caption{Narrow and same types of convolution. The filter has size $3 \times 3$. }
\label{fig:conv_type}
\end{figure}

The classical convolution has smaller output size than input size, \textit{namely}, \textit{narrow} convolution, as shown in Figure~\ref{fig:conv_type}(a). Assume that the input size is $5 \times 5$ and the filter size is $3 \times 3$ with stride 1, the output size is $3 \times 3$ if using narrow convolution. 
In our task, the final output size should be same as the size of the input (\ie{} $I \times J$). For this goal, we employ a special type of convolution, \ie{} \textit{same} convolution (see Figure~\ref{fig:conv_type}(b)), which allows a filter to go outside the border of an input, padding each area outside the border with a zero. 

\noindent\textit{\textbf{Residual Unit.}} 
It is a well-known fact that very deep convolutional networks compromise training effectiveness though the well-known activation function (\eg{} ReLU) and regularization techniques are applied \cite{Ioffe2015,Krizhevsky2012,Nair2010}. 
On the other hand, we still need a very deep network to capture very large citywide dependencies. 
For a typical crowd flow data, assume that the input size is $32 \times 32$, and the kernel size of convolution is fixed to $3 \times 3$, if we want to model citywide dependencies (\ie{}, each node in high-level layer depends on all nodes of the input), it needs more than 15 consecutive convolutional layers. 
To address this issue, we employ residual learning \cite{He2015apa} in our model, which have been demonstrated to be very effective for training super deep neural networks of over-1000 layers. 

In our ST-ResNet (see Figure~\ref{fig:STResNet}), we stack $L$ residual units upon \texttt{Conv1} as follows, 
\begin{equation}\label{eq:res}
\mathbf X_c^{(l+1)} = \mathbf X_c^{(l)} + \mathcal F(\mathbf X_c^{(l)}; \theta_c^{(l)}), l=1,\cdots, L 
\end{equation}
where $\mathcal F$ is the residual function (\ie{} two combinations of ``ReLU + Convolution'', see Figure~\ref{fig:ResUnit}(b)), and $\theta^{(l)}$ includes all learnable parameters in the $l^{th}$ residual unit. We also attempt \textit{Batch Normalization} (BN) \cite{Ioffe2015} that is added before ReLU. 
On top of the $L^{th}$ residual unit, 
we append a convolutional layer (\ie{} \texttt{Conv2} shown in Figure~\ref{fig:STResNet}). 
With 2 convolutions and $L$ residual units, the output of the \textit{closeness} component of Figure~\ref{fig:STResNet} is $\mathbf X_c^{(L+2)}$. 

Likewise, using the above operations, we can construct the \textit{period} and \textit{trend} components of Figure~\ref{fig:STResNet}. 
Assume that there are $l_p$ time intervals from the period fragment and the period is $p$. Therefore, the \textit{period} dependent sequence is $[\mathbf X_{t-{l_p} \cdot p}, \mathbf X_{t-({l_p}-1)\cdot p}, \cdots, \mathbf X_{t-p}]$. 
With the convolutional operation and $L$ residual units like in Eqs.~\ref{eq:conv} and \ref{eq:res}, the output of the \textit{period} component is $\mathbf X_p^{(L+2)}$. 
Meanwhile, the output of the \textit{trend} component is $\mathbf X_q^{(L+2)}$ with the input $[\mathbf X_{t-{l_q} \cdot q}, \mathbf X_{t-({l_q}-1)\cdot q}, \cdots, \mathbf X_{t-q}]$ where $l_q$ is the length of the \textit{trend} dependent sequence and $q$ is the trend span. 
Note that $p$ and $q$ are actually two different types of periods. In the detailed implementation, $p$ is equal to one-day that describes daily periodicity, and $q$ is equal to one-week that reveals the weekly trend. 

\subsection{The Structure of the External Component}
Traffic flows can be affected by many complex external factors, such as weather and event. 
Figure~\ref{fig:holiday} shows that crowd flows during holidays (Chinese Spring Festival) can be significantly different from the flows during normal days. Figure~\ref{fig:weather} shows that heavy rain sharply reduces the crowd flows at Office Area compared to the same day of the latter week. 
Let $E_t$ be the feature vector that represents these external factors at predicted time interval $t$. 
In our implementation, we mainly consider weather, holiday event, and metadata (\ie{} DayOfWeek, Weekday/Weekend). 
The details are introduced in Table~\ref{tab:datasets}. 
To predict flows at time interval $t$, the holiday event and metadata can be directly obtained. However, the weather at future time interval $t$ is unknown. Instead, one can use the forecasting weather at time interval $t$ or the approximate weather at time interval $t-1$. 
Formally, we stack two fully-connected layers upon $E_t$, the first layer can be viewed as an embedding layer for each sub-factor followed by an activation. The second layer is used to map low to high dimensions that have the same shape as $\mathbf X_t$. The output of the 
\textit{external} component of Figure~\ref{fig:STResNet} is denoted as $\mathbf X_{Ext}$ with the parameters $\theta_{Ext}$. 

\begin{figure}[!htbp]
\centering
\subfigure[{Feb 8-14 (red), Feb 15-21 (green), 2016}]{\label{fig:holiday}\includegraphics[width=.32\linewidth]{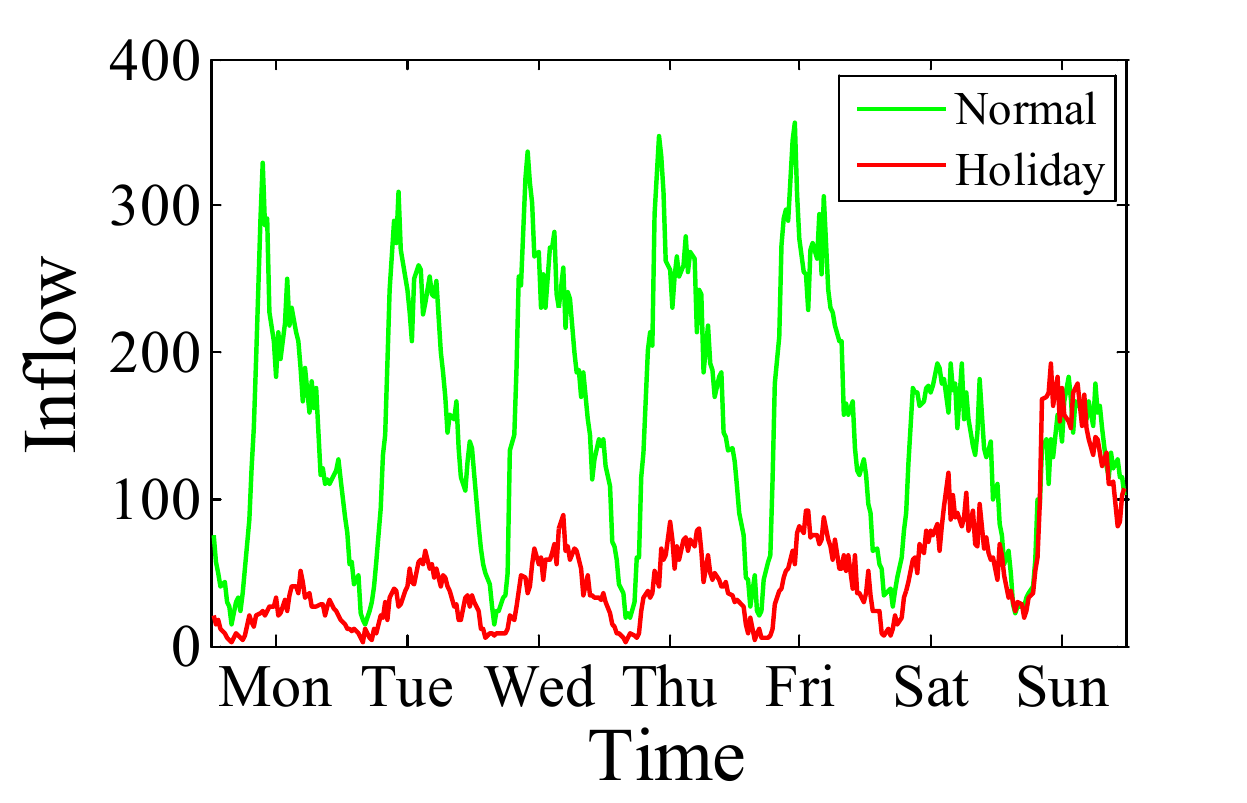}}
\hspace{1em}
\subfigure[{Aug 10-12 (red), Aug 17-19 (green), 2013}]
{\label{fig:weather}\includegraphics[width=.32\linewidth]{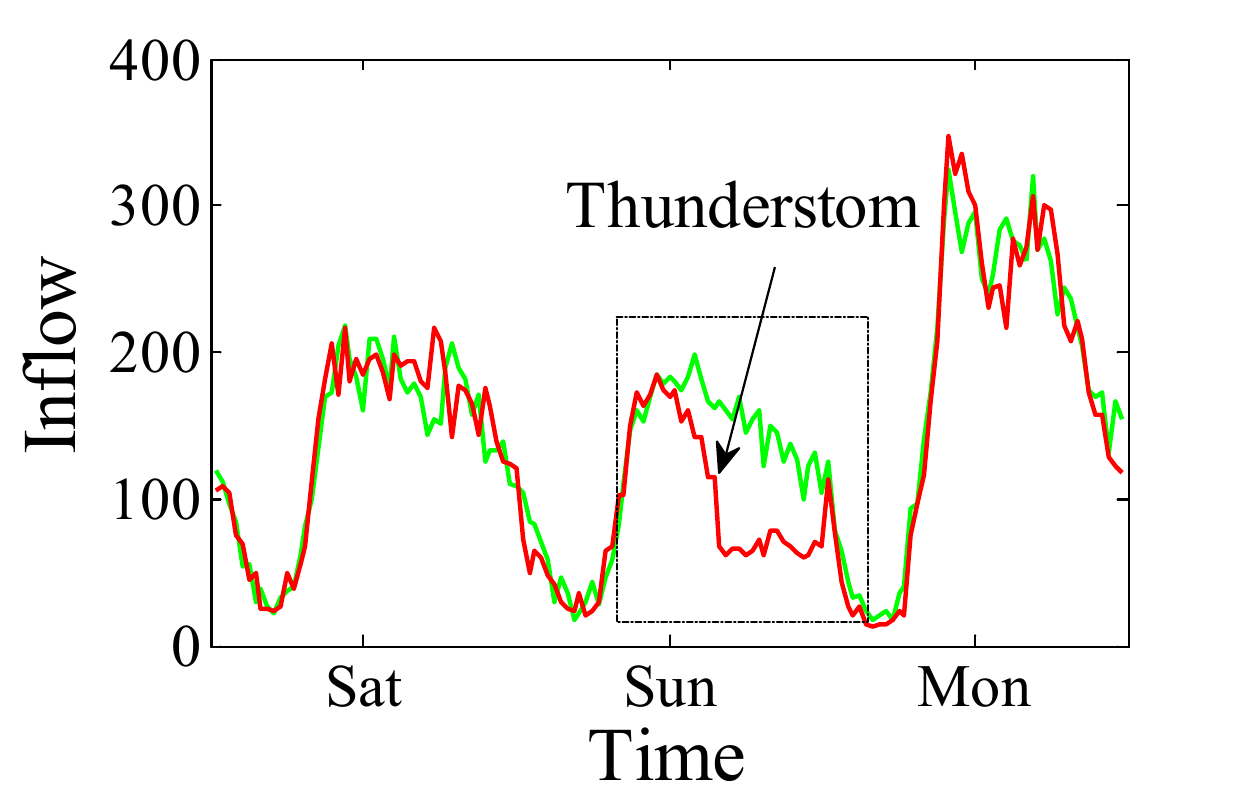}}
\caption{Effects of holidays and weather in Office Area of Beijing (the region is shown in Figure~\ref{fig:gridmap}). }
\label{fig:external}
\end{figure}

\subsection{Fusion}
In this section, we discuss how to fuse four components of Figure~\ref{fig:STResNet}. We first fuse the first three components with a parametric-matrix-based fusion method, which is then further combined with the \textit{external} component. 

Figures~\ref{fig:temporal}(a) and (d) show the ratio curves using Beijing trajectory data presented in Table~\ref{tab:datasets} where $x$-axis is time gap between two time intervals and $y$-axis is the average ratio value between arbitrary two inflows that have the same time gap. 
The curves from two different regions all show an empirical temporal correlation in time series, namely, inflows of recent time intervals are more relevant than ones of distant time intervals, which implies temporal \textit{closeness}. 
The two curves have different shapes, which demonstrates that different regions may have different characteristics of closeness. 
Figures~\ref{fig:temporal}(b) and (e) depict inflows at all time intervals of 7 days. We can see the obvious \textit{daily periodicity} in both regions. In Office Area, the peak values on weekdays are much higher than ones on weekends. Residential Area has similar peak values for both weekdays and weekends. 
Figures~\ref{fig:temporal}(c) and (f) describe inflows at a certain time interval (9:00pm-9:30pm) of Tuesday from March 2015 and June 2015. 
As time goes by, the inflow progressively decreases in Office Area, and increases in Residential Area. 
It shows the different trends in different regions.  
In summary, inflows of two regions are all affected by \textit{closeness}, \textit{period}, and \textit{trend}, but the degrees of influence may be very different. 
We also find the same properties in other regions as well as their outflows. 

\begin{figure}[!htbp]
\centering\label{fig:temporal}\includegraphics[width=.7\linewidth]{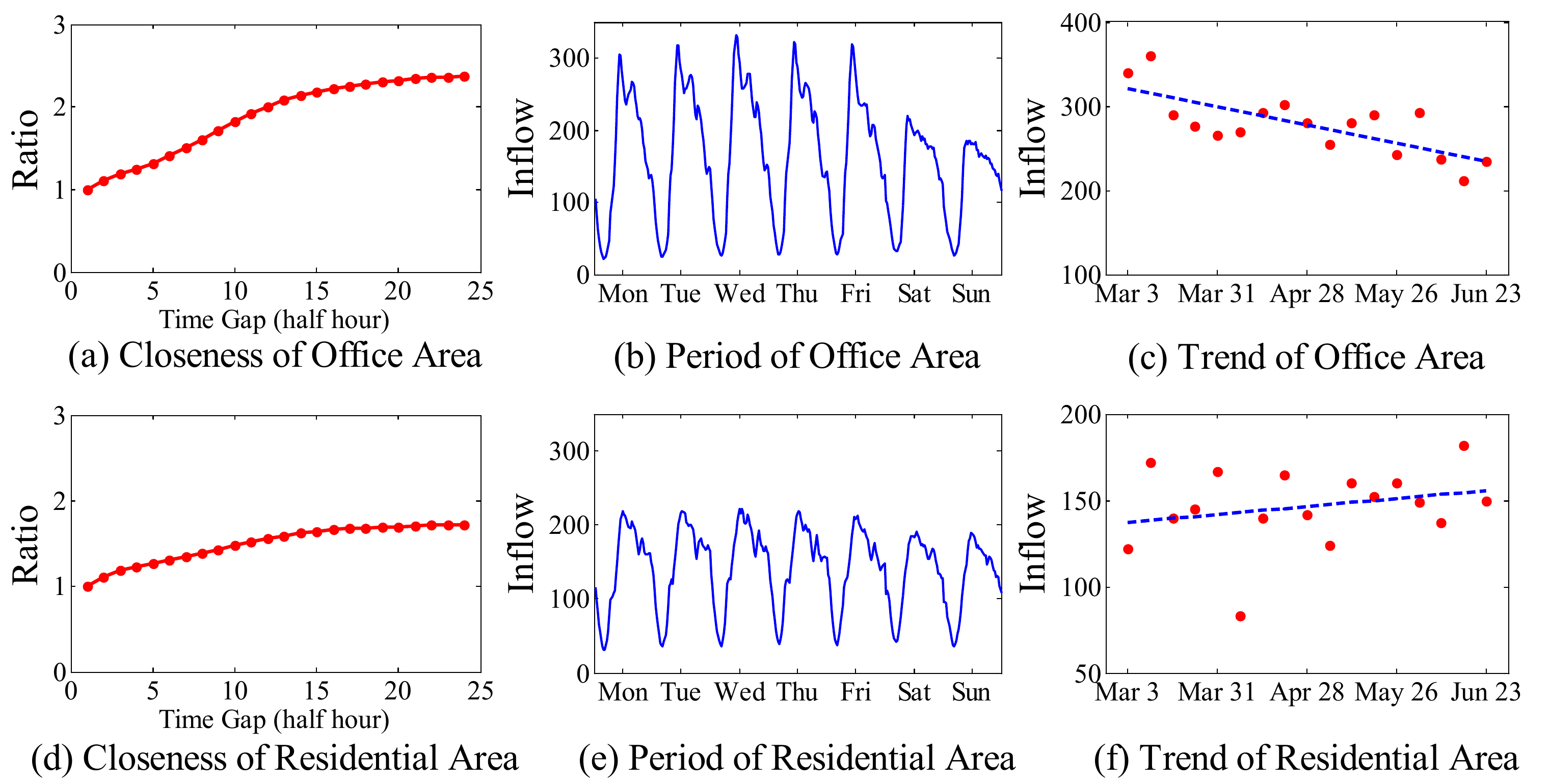}
\caption{Temporal dependencies (Office Area and Residential Area are shown in Figure~\ref{fig:gridmap}). }
\label{fig:temporal}
\end{figure}

Above all, the different regions are all affected by \textit{closeness}, \textit{period} and \textit{trend}, but the degrees of influence may be different. 
Inspired by these observations, we propose a parametric-matrix-based fusion method. 

\noindent{\textbf{\textit{Parametric-matrix-based fusion}}}. 
We fuse the first three components (\ie{} \textit{closeness}, \textit{period}, \textit{trend}) of Figure~\ref{fig:STResNet} as follows
\begin{equation}\label{eq:res_out}
\mathbf X_{Res}=\mathbf W_c \circ \mathbf X_c^{(L+2)} + \mathbf W_p \circ \mathbf X_p^{(L+2)} + \mathbf W_q \circ \mathbf X_q^{(L+2)}
\end{equation}
where $\circ$ is Hadamard product (\ie{} element-wise multiplication), $\mathbf W_c$, $\mathbf W_p$ and $\mathbf W_q$ are the learnable parameters that adjust the degrees affected by closeness, period and trend, respectively. 

\noindent{\textbf{\textit{Fusing the external component}}}. We here directly merge the output of the first three components with that of the \textit{external} component, as shown in Figure~\ref{fig:STResNet}. Finally, the predicted value at the $t^{th}$ time interval, denoted by $\widehat{\mathbf X}_t$, is defined as 
\begin{equation}\label{eq:output}
\widehat{\mathbf X}_t = \tanh (\mathbf X_{Res} + \mathbf X_{Ext})
\end{equation}
where $\tanh$ is a hyperbolic tangent that ensures the output values are between -1 and 1.

Our ST-ResNet can be trained to predict $\mathbf X_t$ from three sequences of flow matrices and external factor features by minimizing mean squared error between the predicted flow matrix and the true flow matrix: 

\begin{equation} \label{eq:loss}
\mathcal L(\theta) = \|\mathbf X_t - \widehat{\mathbf X}_t \| ^2_2
\end{equation} 
where $\theta$ are all learnable parameters in the ST-ResNet. 

\subsection{Algorithms and Optimization}\label{sec:alg}
Algorithm~\ref{Alg:STResNet} outlines the ST-ResNet training process. 
We first construct the training instances from the original sequence data (lines 1-6). 
Then, ST-ResNet is trained via backpropagation and Adam \cite{Kingma2014apa} (lines 7-11). 
\begin{algorithm}[!htbp]\label{Alg:STResNet}\fontsize{9}{9}\selectfont
\KwIn{Historical observations: $\{\mathbf X_0, \cdots, \mathbf X_{n-1}\}$; \\
\qquad\quad external features: $\{E_0, \cdots, E_{n-1}\}$; \\
\qquad\quad lengths of \textit{closeness}, \textit{period}, \textit{trend} sequences: $l_c$, $l_p, l_q;$\\
\qquad\quad peroid: $p$; trend span: $q$.}
\KwOut{ST-ResNet model $\mathcal M$.}
{ 
   \DontPrintSemicolon
	{\tcp*[h]{construct training instances}}\;
	$\mathcal D \longleftarrow \emptyset$\; %
	\For{all available time interval $t (1\leq t \leq n-1)$}
	{
	 	$\mathcal S_c=[\mathbf X_{t-{l_c}}, \mathbf X_{t-({l_c}-1)}, \cdots, \mathbf X_{t-1}]$ \;
	 	$\mathcal S_p=[\mathbf X_{t-{l_p} \cdot p}, \mathbf X_{t-({l_p}-1)\cdot p}, \cdots, \mathbf X_{t-p}]$ \;
	 	$\mathcal S_q=[\mathbf X_{t-{l_q} \cdot q}, \mathbf X_{t-({l_q}-1)\cdot q}, \cdots, \mathbf X_{t-q}]$ \;
	 	\tcp*[h]{$\mathbf X_t$ is the target at time $t$}\;
	 	put an training instance $(\{\mathcal S_c, \mathcal S_p,\mathcal S_q, E_t\},  \mathbf X_t)$ into  $\mathcal D$\;
	}
	{\tcp*[h]{train the model}}\;
    initialize the parameters $\theta$\;
	\Repeat{stopping criteria is met}
	{
	randomly select a batch of instances $\mathcal D_b$ from $\mathcal D$\;
	find $\theta$ by minimizing the objective (\ref{eq:loss}) with $\mathcal D_b$\;
	}
	output the learned ST-ResNet model $\mathcal M$
}
\caption{Training of ST-ResNet}
\end{algorithm}

After training, the learned ST-ResNet model $\mathcal M$ is obtained for the single- or multi-step look-ahead prediction. 
the process of which is summarized in Algorithm~\ref{alg:pred}. 
Some types of external features (\ie{}, weather) used here are different from that in Algorithm~\ref{Alg:STResNet}. 
In the training process, we use the true weather data, which is replaced by the forecasted weather data in Algorithm~\ref{alg:pred}. 

\begin{algorithm}[!htbp]\label{alg:pred}\fontsize{9}{9}\selectfont
\KwIn{
Learned ST-ResNet model: $\mathcal M$;\\
\qquad\quad number of look-ahead steps: $k$; \\
\qquad\quad historical observations: $\{\mathbf X_0, \cdots, \mathbf X_{n-1}\}$; \\
\qquad\quad external features: $\{E_{n}, \cdots, E_{n+k-1}\}$; \\
\qquad\quad lengths of \textit{closeness}, \textit{period}, \textit{trend} sequences: $l_c$, $l_p, l_q;$\\
\qquad\quad peroid: $p$; trend span: $q$. }
{   \DontPrintSemicolon
	$\mathcal X \longleftarrow \{\mathbf X_0, \cdots, \mathbf X_{n-1}\}$ \tcp*[h]{(i.e., $\mathcal X_t = \mathbf X_t, \forall t$)}\;
	\For{$t=n$ \KwTo $n+k-1$}
	{
	 	$\mathcal S_c=[\mathcal X_{t-{l_c}}, \mathcal X_{t-({l_c}-1)}, \cdots, \mathcal X_{t-1}]$\;
	 	$\mathcal S_p=[\mathcal X_{t-{l_p} \cdot p}, \mathcal X_{t-({l_p}-1)\cdot p}, \cdots, \mathcal X_{t-p}]$ \;
	 	$\mathcal S_q=[\mathcal X_{t-{l_q} \cdot q}, \mathcal X_{t-({l_q}-1)\cdot q}, \cdots, \mathcal X_{t-q}]$ \;
	 	$\widehat{\mathbf X}_t \longleftarrow$ $\mathcal M(\mathcal S_c, \mathcal S_p,\mathcal S_q, E_t)$\;
	 	put $\widehat{\mathbf X}_t$ into $\mathcal X$, \ie{}, $\mathcal X_t = \widehat{\mathbf X}_t$\;
	}
	output $\{\widehat{\mathbf X}_n, \cdots, \widehat{\mathbf X}_{n+k-1}\}$
}
\caption{Multi-step Ahead Prediction Using ST-ResNet}
\end{algorithm}

\section{Experiments}\label{sec:expt}
In this section, we evaluate our ST-ResNet on two types of crowd flows in Beijing and NYC against 9 baselines. 
\subsection{Settings}\label{sec:setting}
\noindent\textbf{Datasets.} We use two different sets of data as shown in Table~\ref{tab:datasets}. Each dataset contains two sub-datasets: trajectories and weather, as detailed as follows. 

\begin{itemize}
\item \textbf{TaxiBJ}: Trajectoriy data is the taxicab GPS data and meteorology data in Beijing from four time intervals: 1st Jul. 2013 - 30th Otc. 2013, 1st Mar. 2014 - 30th Jun. 2014, 1st Mar. 2015 - 30th Jun. 2015, 1st Nov. 2015 - 10th Apr. 2016. 
Using Definition~\ref{def:flow}, we obtain two types of crowd flows. 
We choose data from the last four weeks as the testing data, and all data before that as training data. 
\item \textbf{BikeNYC}: Trajectory data is taken from the NYC Bike system in 2014, from Apr. 1st to Sept. 30th. Trip data includes: trip duration, starting and ending station IDs, and start and end times. Among the data, the last 10 days are chosen as testing data, and the others as training data. 
\end{itemize}

\begin{table}[!htbp]%
\tabcolsep 0pt 
\begin{center}
\caption{Datasets (holidays include adjacent weekends).}\label{tab:datasets}
\begin{tabular}{ccc}
\hline
\hline
\textbf{Dataset} & \textbf{TaxiBJ} & \textbf{BikeNYC} \\
\hline
Data type & Taxi GPS & Bike rent\\
Location & Beijing & New York \\
\cline{2-3}
\multirow{4}*{Time Span} & 7/1/2013 - 10/30/2013 & \\
&3/1/2014 - 6/30/2014&4/1/2014 - 9/30/2014\\
&3/1/2015 - 6/30/2015&\\
&11/1/2015 - 4/10/2016&\\
\cline{2-3}
Time interval & 30 minutes & 1 hour \\
Gird map size & (32, 32) & (16, 8) \\
\hline
\multicolumn{3}{c}{\textbf{Trajectory data}} \\
Average sampling rate (s) & $\sim$ 60 & $\setminus$ \\
\# taxis/bikes & 34,000+ & 6,800+ \\
\# available time interval & 22,459  & 4,392 \\
\hline
\multicolumn{3}{c}{\textbf{External factors (holidays and meteorology)}} \\
\# holidays & 41 & 20 \\
Weather conditions & 16 types (\eg{}, Sunny, Rainy) & $\setminus$ \\
Temperature / $^\circ$C  & $[-24.6, 41.0]$ & $\setminus$ \\ %
Wind speed / mph & $[0, 48.6]$ & $\setminus$ \\
\hline
\hline
\end{tabular}
\end{center}
\end{table}

\noindent\textbf{Baselines}. We compare our ST-ResNet with the following 9 baselines: %
\begin{itemize}
\item \textbf{HA}: We predict inflow and outflow of crowds by the average value of historical inflow and outflow in the corresponding periods, \eg{}, 9:00am-9:30am on Tuesday, its corresponding periods are all historical time intervals from 9:00am to 9:30am on all historical Tuesdays. 
\item \textbf{ARIMA}: Auto-Regressive Integrated Moving Average (ARIMA) is a well-known model for understanding and predicting future values in a time series. 
\item \textbf{SARIMA}: Seasonal ARIMA. Beyond ARIMA, SARIMA also considers the seasonal terms, capable of both learning closeness and periodic dependencies. 
\item \textbf{VAR}: Vector Auto-Regressive (VAR) is a more advanced spatio-temporal model, which can capture the pairwise relationships among all flows, and has heavy computational costs due to the large number of parameters. 
\item \textbf{ST-ANN}: It first extracts spatial (nearby 8 regions' values) and temporal (8 previous time intervals) features, then fed into an artificial neural network. %
\item \textbf{DeepST} \cite{Zhang2016}: a deep neural network (DNN)-based prediction model for spatio-temporal data, which shows state-of-the-art results on the crowd flow prediction. %
\item \textbf{RNN} \cite{bengiodeep}: recurrent neural network (RNN), a deep learning model, which can capture temporal dependencies. Formally, RNN can train on sequences with the arbitrary length. In our experiment, we fix the length of input sequence as one of $\{3, 6, 12, 24, 48, 336\}$. Taking 48 as example, the dependent input sequence is just a one-day data if the interval time is equal to 30 minutes. Therefore, we have 6 RNN variants, including RNN-3, RNN-6, RNN-12, RNN-24, RNN-48, and RNN-336. 
\item \textbf{LSTM} \cite{hochreiter1997long}: Long-short-term-memory network (LSTM), a special kind of RNN, capable of learning long-term temporal dependencies. Being same as the setting of RNN, we conduct the experiments on 6 LSTM variants, \ie{} LSTM-3, LSTM-6, LSTM-12, LSTM-24, LSTM-48, and LSTM-336. 
\item \textbf{GRU} \cite{cho2014learning}: Gated-recurrent-unit network, a new kind of RNN, can be used to capture long-term temporal dependencies. Being same as the setting of RNN, the following GRU variants are selected as the baselines: GRU-3, GRU-6, GRU-12, GRU-24, GRU-48, and GRU-336. 
\end{itemize}

\noindent\textbf{Preprocessing.} In the output of the ST-ResNet, we use $\tanh$ as our final activation (see Eq.~\ref{eq:output}), whose range is between -1 and 1. Here, we use the Min-Max normalization method to scale the data into the range $[-1, 1]$. In the evaluation, we re-scale the predicted value back to the normal values, compared with the groundtruth. For external factors, we use one-hot coding to transform metadata (\ie{}, DayOfWeek, Weekend/Weekday), holidays and weather conditions into binary vectors, and use Min-Max normalization to scale the Temperature and Wind speed into the range $[0, 1]$. 

\noindent\textbf{Hyperparameters.} The learnable parameters are initialized using a uniform distribution with the default parameter in Keras \cite{Chollet2015}.  
The convolutions of \texttt{Conv1} and all residual units use 64 filters of size $3\times 3$, and \texttt{Conv2} uses a convolution with 2 filters of size  $3\times 3$. For example, a 4-residual-unit of ST-ResNet consists of \texttt{Conv1}, 4 residual unit, and \texttt{Conv2}. See Table~\ref{tab:arch} for the details. 
The Adam \cite{Kingma2014apa} is used for optimization, and the batch size is 32. The number of residual units is set as 12 for the dataset TaxiBJ, and 4 for BikeNYC. 
There are 5 extra hyperparamers in our ST-ResNet, of which $p$ and $q$ are empirically fixed to one-day and one-week, respectively. 
For lengths of the three dependent sequences, we set them as: $l_c \in \{1,2,3,4,5\}, l_p \in \{1,2,3,4\}, l_q \in \{1,2,3,4\}$. 
We select 90\% of the training data for training each model, and the remaining 10\% is chosen as the validation set, which is used to early-stop our training algorithm for each model based on the best validation score. 
Afterwards, we continue to train the model on the full training data for a fixed number of epochs (\eg{}, 10, 100 epochs). 

\begin{table}[!htbp]
\tabcolsep 8pt
\begin{center}
\caption{Details of convolutions and residual units}
\label{tab:arch} 
\begin{tabular}{c|c|c|c|c}
\hline
\hline
layer name & output size & closeness & period & trend \\
\hline
\texttt{Conv1} &  $32\times 32$ & $3\times 3, 64 $& $3\times 3, 64$ & $3\times 3, 64$\\
ResUnit 1 &  $32\times 32$ & $\left[\begin{array}{c} 3\times 3, 64  \\  3\times 3, 64  \\ \end{array} \right] \times 2 $ & $\left[\begin{array}{c} 3\times 3, 64  \\  3\times 3, 64  \\ \end{array} \right] \times 2 $ & $\left[\begin{array}{c} 3\times 3, 64  \\  3\times 3, 64  \\ \end{array} \right] \times 2 $\\
ResUnit 2 &  $32\times 32$ & $\left[\begin{array}{c} 3\times 3, 64  \\  3\times 3, 64  \\ \end{array} \right] \times 2 $ & $\left[\begin{array}{c} 3\times 3, 64  \\  3\times 3, 64  \\ \end{array} \right] \times 2 $ & $\left[\begin{array}{c} 3\times 3, 64  \\  3\times 3, 64  \\ \end{array} \right] \times 2 $\\
ResUnit 3 &  $32\times 32$ & $\left[\begin{array}{c} 3\times 3, 64  \\  3\times 3, 64  \\ \end{array} \right] \times 2 $ & $\left[\begin{array}{c} 3\times 3, 64  \\  3\times 3, 64  \\ \end{array} \right] \times 2 $ & $\left[\begin{array}{c} 3\times 3, 64  \\  3\times 3, 64  \\ \end{array} \right] \times 2 $\\
ResUnit 4 &  $32\times 32$ & $\left[\begin{array}{c} 3\times 3, 64  \\  3\times 3, 64  \\ \end{array} \right] \times 2 $ & $\left[\begin{array}{c} 3\times 3, 64  \\  3\times 3, 64  \\ \end{array} \right] \times 2 $ & $\left[\begin{array}{c} 3\times 3, 64  \\  3\times 3, 64  \\ \end{array} \right] \times 2 $\\
\texttt{Conv2} &  $32\times 32$ & $3\times 3, 2 $& $3\times 3, 2$ & $3\times 3, 2$\\
\hline
\hline
\end{tabular}
\end{center}
\end{table}

\noindent\textbf{Evaluation Metric}: We measure our method by Root Mean Square Error (RMSE)\footnote{The smaller the better.} as
\begin{equation}
RMSE = \sqrt { \frac{1}{z} \sum_i (x_i - \hat x_i)^2 } \nonumber
\end{equation}
where $x$ and $\hat x$ are the available ground truth and the corresponding predicted value, respectively; $z$ is the number of all available ground truths. 

Experiments are mainly run on a GPU server, whose detailed information is shown in Table~\ref{tab:GPU}. The python libraries, including Theano \cite{TDT2016ae} and Keras \cite{Chollet2015}, are used to build our models. 

\begin{table}[!htbp]
\tabcolsep 8pt
\begin{center}
\caption{Experimental environment}
\label{tab:GPU} 
\begin{tabular}{c|c}
\hline
\hline
OS & Windows Server 2012 R2\\
Memory & 256GB \\
CPU & Intel(R) Xeon(R) CPU E5-2680 v2 @ 2.80GHz \\
GPU & Tesla K40m \\
Number of GPU cards & 4 \\
\hline
CUDA version & 8.0\\
cuDNN version & 8.0\\
Keras version & 1.1.1\\
Theano version & 0.9.0dev \\
\hline
\hline
\end{tabular}
\end{center}
\end{table}

\subsection{Evaluation of Single-step Ahead Prediction}\label{sec:expt:single}
\begin{figure}[!htbp]
\centering
\subfigure[TaxiBJ]{\label{fig:one_step:TaxiBJ}\includegraphics[clip, width=.49\linewidth]{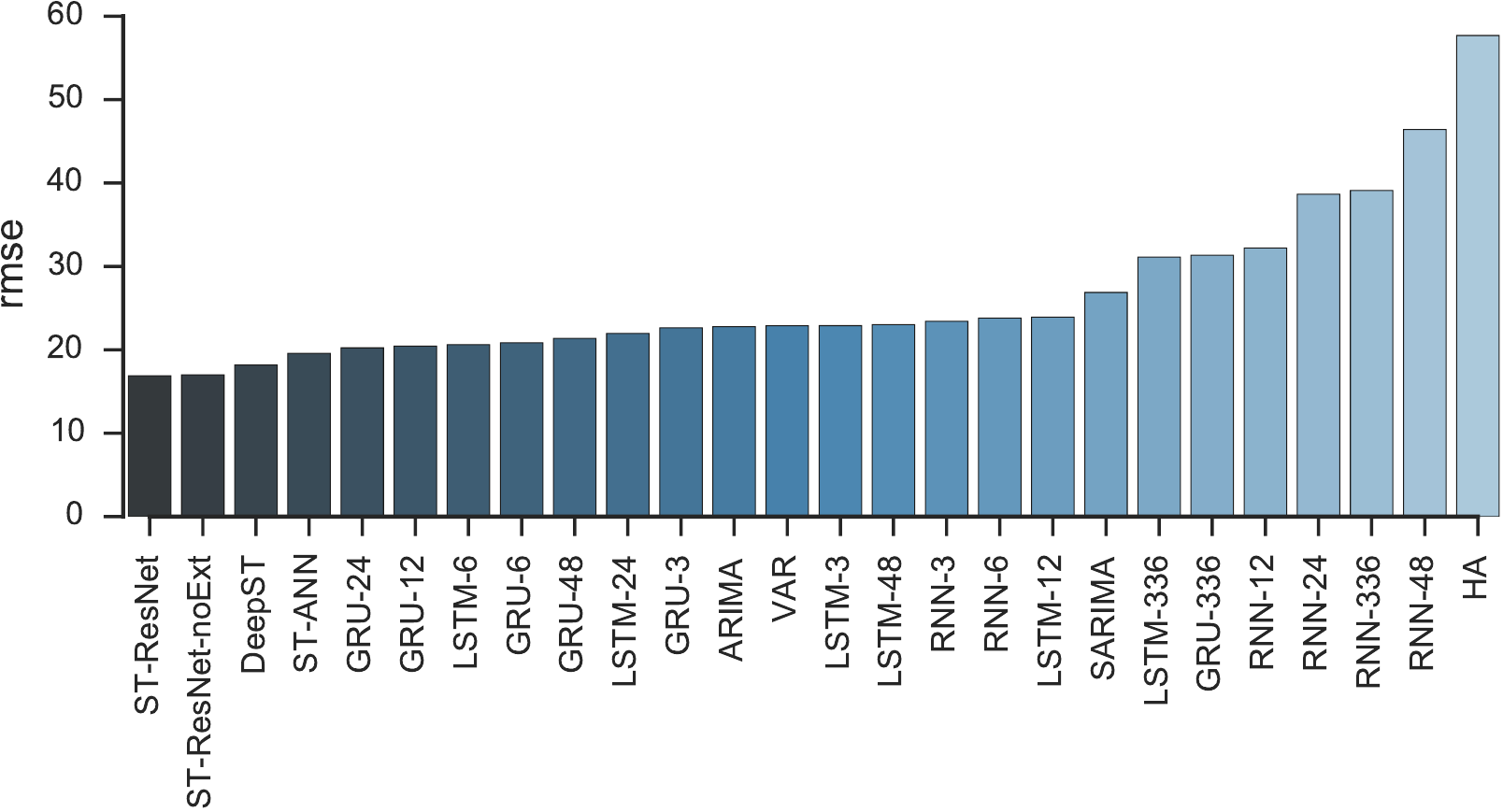}}
\subfigure[BikeNYC]{\label{fig:one_step:BikeNYC}\includegraphics[clip, width=.49\linewidth]{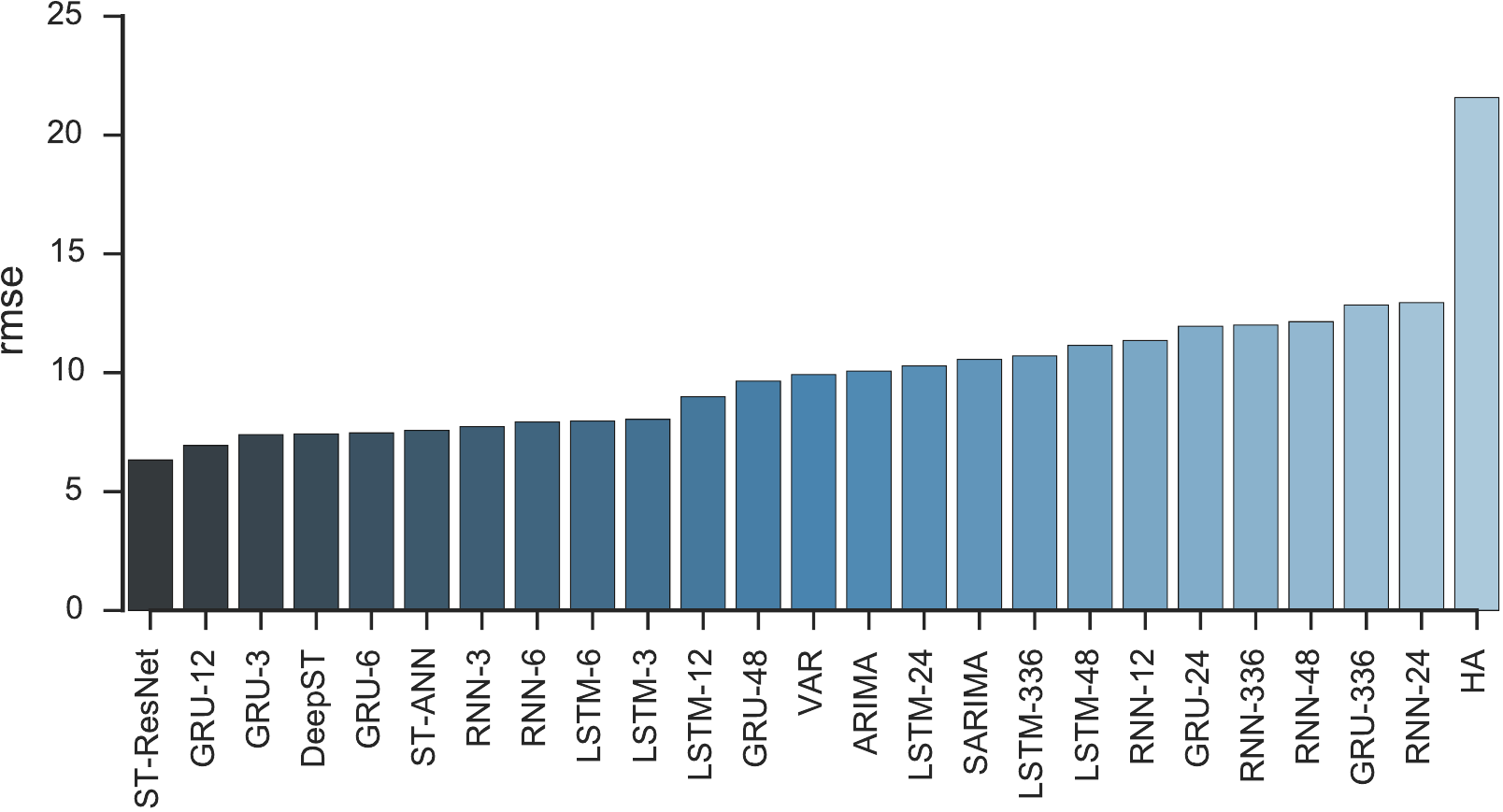}}
\caption{Model ranking on TaxiBJ and BikeNYC. The smaller the better. }
\label{fig:one_step}
\end{figure}

In this section, we evaluate the single-step ahead prediction, \textit{namely}, predicting the crowd flows at time $t$ using the historical observations. 
Table~\ref{tab:prediction} shows the RMSE of all methods on both TaxiBJ and BikeNYC. 
Our ST-ResNet consistently and significantly outperforms all baselines. 
Specifically, the results on TaxiBJ demonstrates that ST-ResNet (with 12 residual units) is relatively $26\%$ better than ARIMA, $37\%$ better than SARIMA, $26\%$ better than VAR, $14\%$ better than ST-ANN, $7\%$ better than DeepST, $28\%$ to $64\%$ better than RNN, $18.1\%$ to $45.7\%$ better than LSTM, $17.4\%$ to $46.1\%$ better than GRU. 
ST-ResNet-noExt is a degraded version of ST-ResNet that does not consider the \textit{external} factors (\eg{} meteorology data). We can see that ST-ResNet-noExt is slightly worse than ST-ResNet, pointing out external factors are always beneficial. 
DeepST exploits spatio-temporal CNNs and is clearly better than other baselines. 
While both ST-ANN and VAR use spatial/temporal information and relationships among flows, they are worse than DeepST because they only consider the \textit{near} spatial information and \textit{recent} temporal information. 
Among the temporal models, GRU and LSTM have similar RMSE, and outperform RNN in average because GRU and LSTM both can capture long-term temporal dependencies. However, GRU-336 and LSTM-336 have very bad performance as well as RNN-336, which demonstrates RNN-based models cannot capture very long-term dependencies (\ie{} \textit{period} and \textit{trend}). 
Intuitively, we rank all of these models, as shown in Figure~\ref{fig:one_step:TaxiBJ}. 

Being different from TaxiBJ, BikeNYC consists of two different types of crowd flows, including new-flow and end-flow \cite{Hoang2016}. 
We here adopt a total of 4-residual-unit ST-ResNet, and consider the metadata as external features like DeepST \cite{Zhang2016}. 
ST-ResNet has relatively from $9\%$ up to $71\%$ lower RMSE than these baselines, demonstrating that our proposed model has good generalization performance on other flow prediction tasks. Figure~\ref{fig:one_step:BikeNYC} depicts the ranking of these models. 

\begin{table}[!htbp]
\tabcolsep 8pt
\begin{center}
\caption{Comparisons with baselines on TaxiBJ and BikeNYC. The results of ARIMA, SARIMA, VAR and DeepST on BikeNYC are taken from \cite{Zhang2016}. }
\label{tab:prediction} 
\begin{tabular}{l|r|r}
\hline
\hline
&\multicolumn{2}{c}{RMSE}\\
\hline
Model&TaxiBJ&BikeNYC\\
\hline
HA&57.69&21.58\\
ARIMA&22.78&10.07\\
SARIMA&26.88&10.56\\
VAR&22.88&9.92\\
ST-ANN&19.57&7.57\\
DeepST&18.18&7.43\\
\hline
RNN-3&23.42&7.73\\
RNN-6&23.80&7.93\\
RNN-12&32.21&11.36\\
RNN-24&38.66&12.95\\
RNN-48&46.41&12.15\\
RNN-336&39.10&12.01\\
\hline
LSTM-3&22.90&8.04\\
LSTM-6&20.62&7.97\\
LSTM-12&23.93&8.99\\
LSTM-24&21.97&10.29\\
LSTM-48&23.02&11.15\\
LSTM-336&31.13&10.71\\
\hline
GRU-3&22.63&7.40\\
GRU-6&20.85&7.47\\
GRU-12&20.46&6.94\\
GRU-24&20.24&11.96\\
GRU-48&21.37&9.65\\
GRU-336&31.34&12.85\\
\hline
ST-ResNet   &  \textbf{16.89} (12 residual units) & \textbf{6.33} (4 residual units)\\
ST-ResNet-noExt & 17.00 (12 residual units) & $\setminus$ \\
\hline
\hline
\end{tabular}
\end{center}
\end{table}

\subsection{Results of Different ST-ResNet Variants}\label{sec:expt:param}
We here present the results of different  ST-ResNet variants, including changing network configurations, network depth, and different components used. 
\subsubsection{Impage of different network configurations}
\begin{figure}[!htbp]
\centering
\includegraphics[clip,width=.8\linewidth]{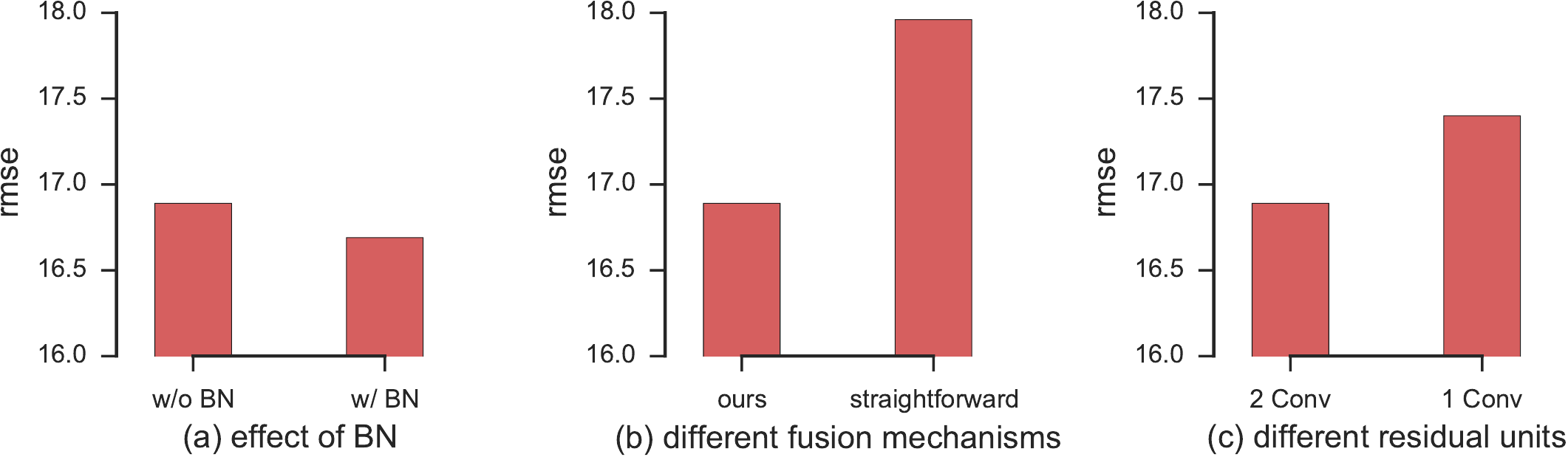}
\vspace{6pt}
\caption{Results of different configurations}
\label{fig:conf}
\end{figure}

Figure~\ref{fig:conf} shows the results of different network configurations. 
The same hyper-parameters: $l_c=3$, $l_p=1$, $l_q=1$, $\textit{number of residual unit} = 12$. 

\begin{itemize}
\item \textit{Effect of batch normalization (BN)}: We attempt to adopt BN into each residual unit, finding that the RMSE slightly improves in single-step ahead prediction, as shown in Figure~\ref{fig:conf}(a).
\item \textit{Effect of parametric-matrix-based fusion}: We use a parametric-matrix-based fusion mechanism  (see Eq.~\ref{eq:res_out})  to fuse temporal \textit{closeness}, \textit{period} and \textit{trend} components. Simply, one also can employ a straightforward method for fusing, \ie{}, $\mathbf X_c^{(L+2)} + \mathbf X_p^{(L+2)} + \mathbf X_q^{(L+2)}$. Figure~\ref{fig:conf}(b) shows that ours is significantly better than the straightforward method, demonstrating the effectiveness of our proposed parametric-matrix-based fusion. 
\item \textit{Internal structure of residual unit}: The proposed residual unit includes 2 convolutions. We here test the performance different setting in the residual unit. From Figure~\ref{fig:conf}(c), we observe that the model using 2 convolutions are better than using 1 convolution. 
\end{itemize}

\subsubsection{Impact of network depth}
\begin{figure}[!htbp]
\centering
\includegraphics[clip,width=.5\linewidth]{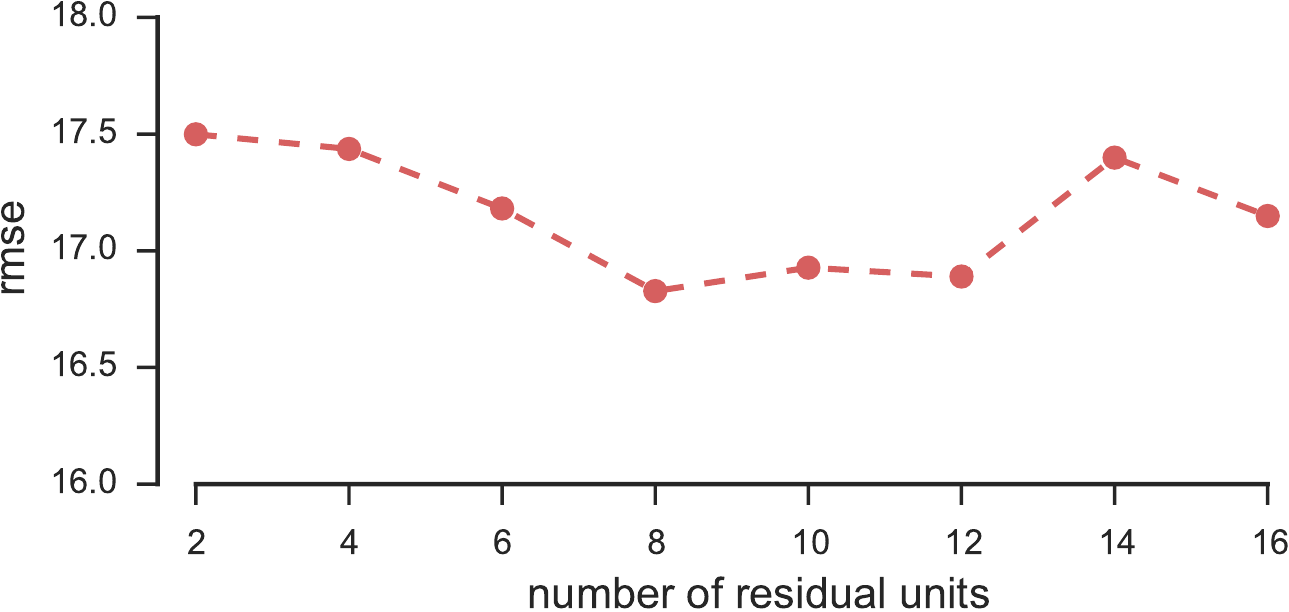}
\vspace{6pt}
\caption{Impact of network depth}
\label{fig:network_depth}
\end{figure}

Figure~\ref{fig:network_depth} presents the impact of network depth. 
As the network goes deeper (\ie{} the number of residual units increases), the RMSE of the model first decreases and then increases, demonstrating that the deeper network often has a better result because it can capture not only near spatial dependence but also distant one. However, when the network is very deep (\eg{} number of residual unit $\geq 14$), training becomes much difficult.   
\vspace{10pt}

\subsubsection{Impact of filter size and number}
\begin{figure}[!htbp]
\centering
\subfigure[filter size]{\label{fig:filter_size}\includegraphics[width=.3\linewidth]{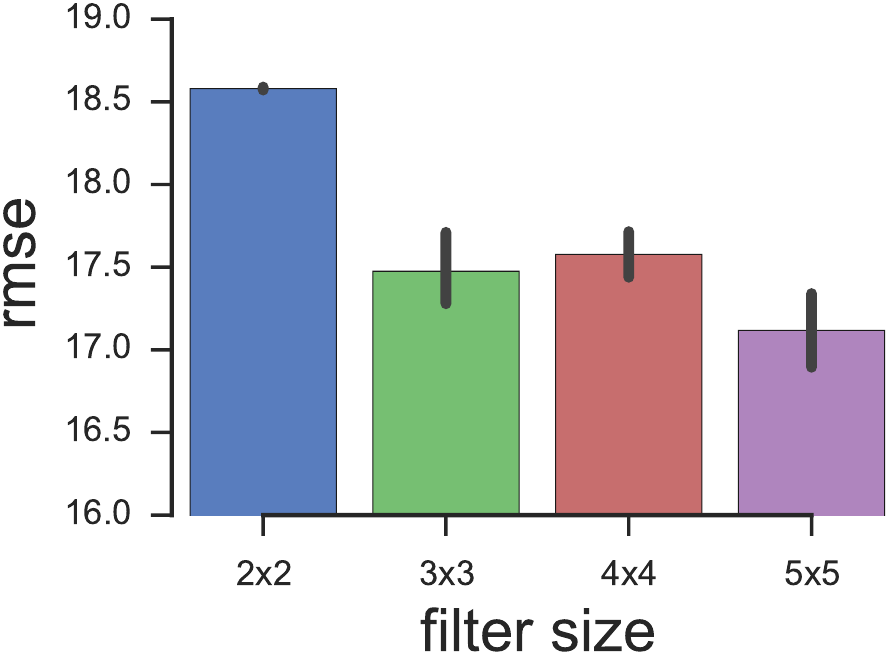}}
\subfigure[filter number]{\label{fig:nb_filter}\includegraphics[width=.3\linewidth]{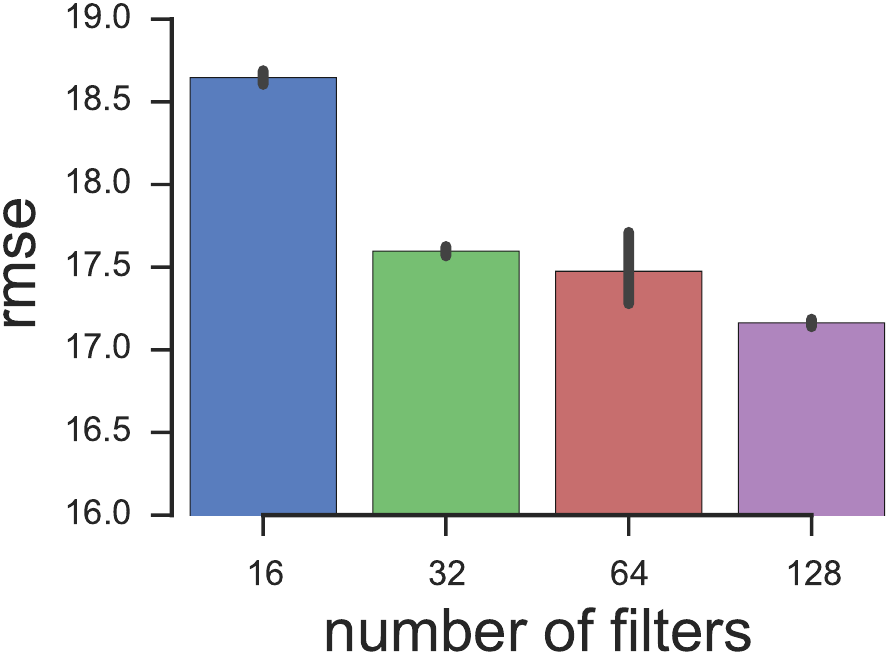}}
\caption{Impact of filter size and number. }
\label{fig:filter}
\end{figure}

The receptive field of a convolution is determined by the size of the filter used. 
We here change the size of the filter from $2\times 2$ to $5\times 5$. 
Figure~\ref{fig:filter_size} shows that the larger filter size has the lower RMSE, demonstrating larger receptive field has better ability to model spatial dependency. 
From Figure~\ref{fig:nb_filter}, we can observe that more filters better result. 

\subsubsection{Impact of temporal closeness, period, trend}
\begin{figure}[!htbp]
\centering
\includegraphics[clip,width=.8\linewidth]{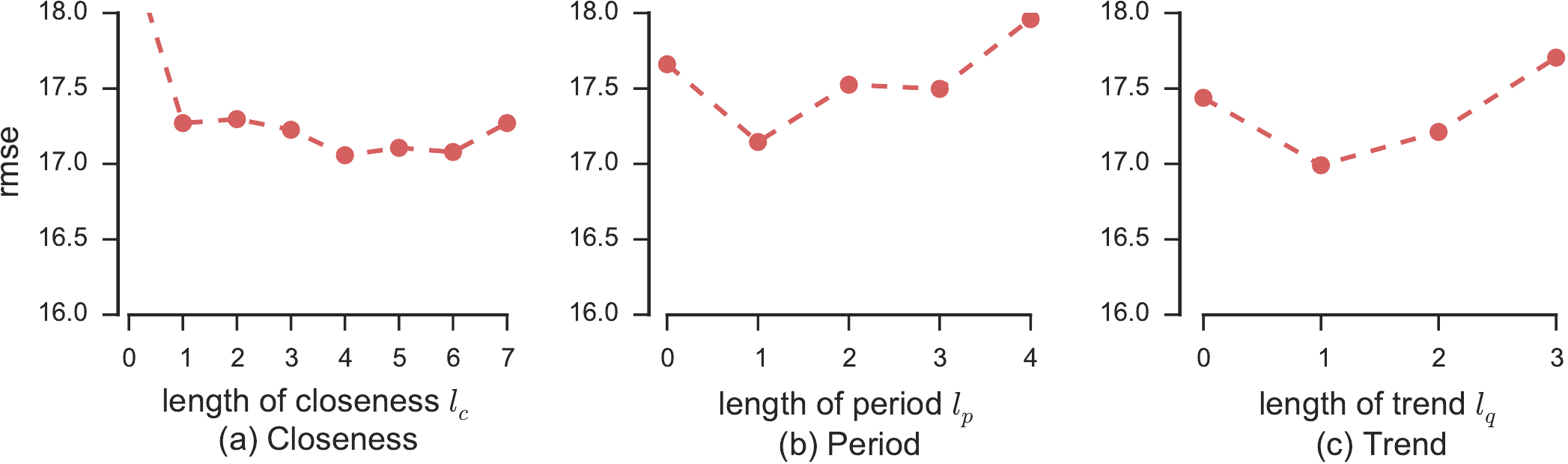}
\caption{Impact of temporal \textit{closeness}, \textit{period}, \textit{trend}}
\label{fig:CTP}
\end{figure}

We here verify the impact of temporal \textit{closeness},\textit{period}, \textit{trend} components on TaxiBJ, as shown in Figure~\ref{fig:CTP}. 
Figure~\ref{fig:CTP}(a) shows the effect of temporal \textit{closeness} where we fix $l_p=1$ and $l_q=1$ but change $l_c$. 
For example, $l_c=0$ means that we does not employ the \textit{closeness} component, resulting in a very bad RMSE: $35.04$. 
We can observe that RMSE first decreases and then increases as the length of closeness increases, indicating that $l_c=4$ has the best performance. 
Figure~\ref{fig:CTP}(b) depicts the effect of \textit{period} where we set $l_c$ as 3 and $l_q$ as 1 but change $l_p$. We can see that $l_p=1$ has the best RMSE. The model without the \textit{period} component (\ie{} $l_p=0$) is worse than the model with $l_p=2, 3$, but better than the $l_p=4$ model, meaning that short-range periods are always beneficial, and long-range periods may be hard to model or not helpful. 
Figure~\ref{fig:CTP}(c) presents the effect of \textit{trend} where $l_c$ and $l_p$ are fixed to 3 and 2, respectively. 
We change $l_q$ from 0 to 3. The curve points that the $l_q=1$ model outperforms others. Similar to \textit{period}, it is better to employ the \textit{trend} component, but long-range trend may be not easy to capture or useless.  

\vspace{10pt}

\begin{figure}[!htbp]
\centering
\includegraphics[clip,width=.8\linewidth]{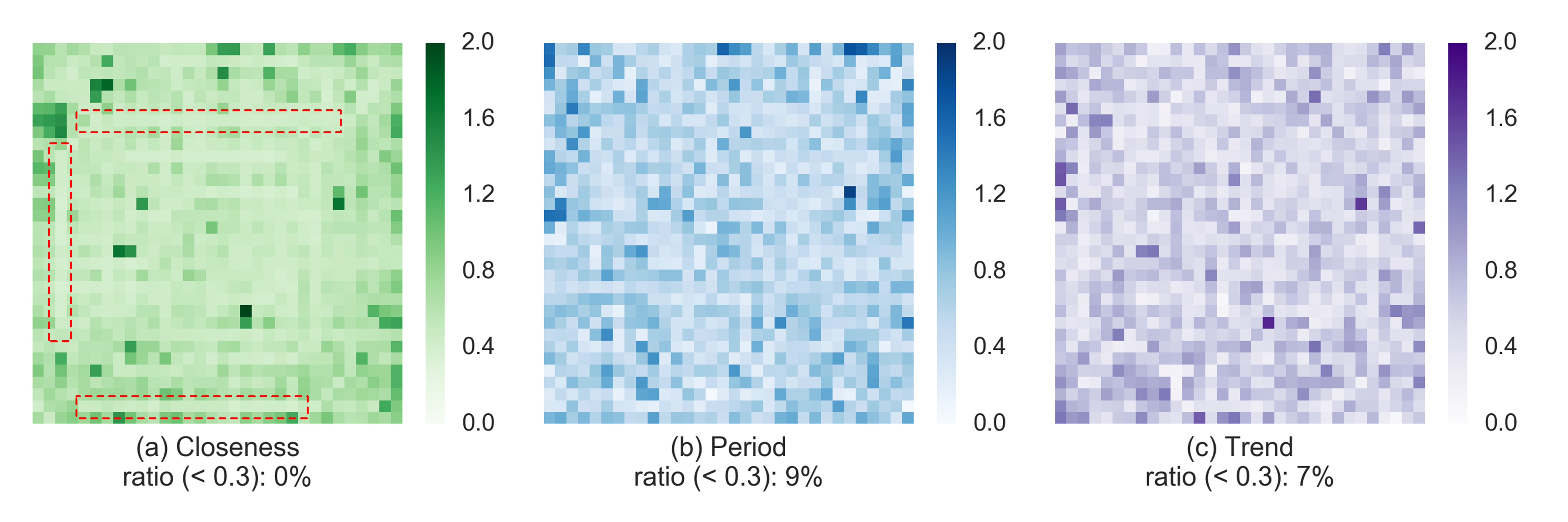}
\caption{Visualization of parameters}
\label{fig:fusion_param}
\end{figure}

To better understand the temporal \textit{closeness}, \textit{period} and \textit{trend}, 
we here visualize the parameters of the parametric-matrix-based fusion layer, which is capable of learning different temporal influence degrees for each region of a city, as shown in Figure~\ref{fig:fusion_param}. %
Each element in each sub-figure denotes a learned parameter of a certain region that reflects the influence degree by \textit{closeness}, \textit{period}, or \textit{trend}. 
We here set a threshold (\eg{}, $0.3$) to see the temporal properties of the whole city. 
Given a fixed threshold $0.3$, we observe that the ratio (the number of regions whose parametric value is less than $0.3$) of the \textit{closeness} is 0, demonstrating all of regions in the city have a more or less \textit{closeness}. 
The ratio of the \textit{period} shows that there are $9\%$ regions only have very weak periodic patterns. Likewise, Figure~\ref{fig:fusion_param}(c) depicts that $7\%$ regions do not have temporal \textit{trend}. 
From Figure~\ref{fig:fusion_param}(a), we find that the \textit{closeness} of some main-road-related regions (red dashed frame) is not obvious. One reason is that the crowd flows in these regions can be predicted using \textit{period} or/and \textit{trend}, adding slight \textit{closeness}.

\subsection{Evaluation of Multi-step Ahead Prediction}\label{sec:expt:multi}
\begin{figure}[!htbp]
\centering
\includegraphics[clip,width=.9\linewidth]{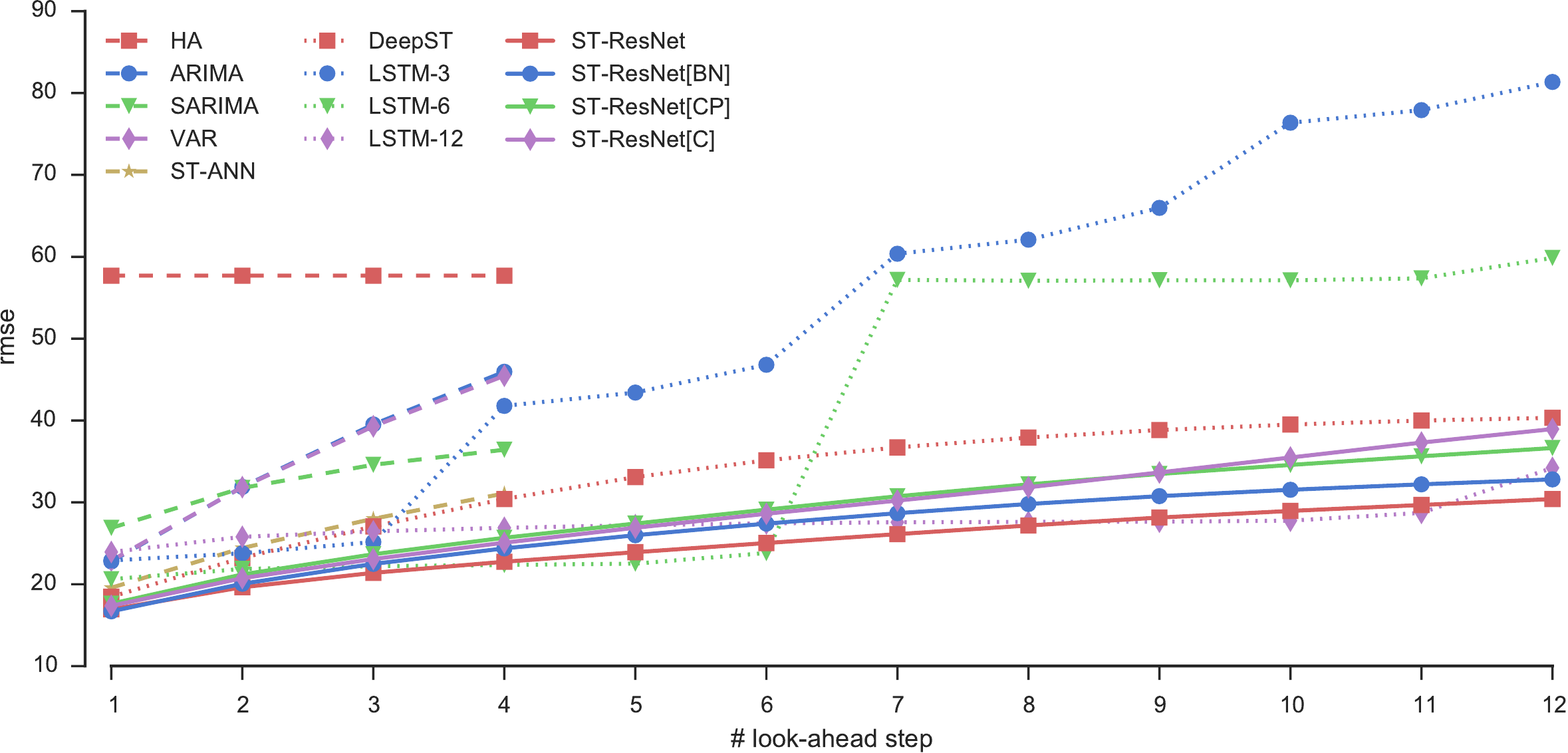}
\caption{Multi-step ahead prediction}
\label{fig:multistep}
\end{figure}

According to Algorithm~\ref{alg:pred}, we can use historical observations and the recent predicted ones to forecast the crowd flows in subsequent time intervals which is referreed to multi-step ahead prediction. 
Figure~\ref{fig:multistep} shows multi-step prediction results of 13 different models on TaxiBJ. Among these models, ST-ResNet[BN], ST-ResNet[CP], and ST-ResNet[C] are three variants of ST-ResNet (12 residual units), of which ST-ResNet[BN] employs BN in all residual units, ST-ResNet[CP] does not employ the \textit{trend} component but three others, ST-ResNet[C] only uses the \textit{closeness} and \textit{external} components. LSTM-3, LSTM-6 and LSTM-12 are three variants of LSTM (see details in Section~\ref{sec:setting}). 
In real-world applications, forecasting the crowd flows in the near future (\eg{} future 2 hours) is much more important. 
From the results of 4-step ahead prediction\footnote{4-step ahead prediction on TaxiBJ means predicting the crowd flows in next 2 hours}, we find our ST-ResNet performs best though ST-ResNet[BN] is better in the single-step ahead prediction, showing in Figure~\ref{fig:conf}(a). 
From the curves of ST-ResNet, ST-ResNet[C] and ST-ResNet[CP], we observe that ST-ResNet is significantly best, demonstrating the \textit{period} and \textit{trend} are very important in the multi-step ahead prediction. 
We observe that LSTM-12 is better than ST-ResNet when the number of the look-ahead steps is greater than 8. 
The reason may be that LSTM-12 reads the past 12 observations to predict, however, our ST-ResNet only takes past recent 3 observations as the input of the \textit{closeness} component. 

\subsection{Efficiency and Resources}\label{sec:expt:cloud}
We test the efficiency on two different virtual machines in the cloud (\ie{} Microsoft Azure). 
As introducing in Section~\ref{sec:cloud}, there are four main steps to predict crowd flows for each region of a city: 
(1) pulling trajectories from redis; (2) converting trajectories into crowd flow data; (3) predicting the crowd flows in near future; (4) pushing results into redis. 
We also report the time consumed by above four steps. 
Totally, A2 standard VM finishes the whole predicting process in 18.56 seconds. It takes 10.93 seconds on D4 standard VM, which is more powerful but expensive. One can choose \textit{A2 standard} because it only costs ~20\% money but achieves more than 50\% performance. 

\begin{table}[!htbp]
\tabcolsep 8pt
\begin{center}
\caption{Configuration of virtual machines and performance}
\label{tab:azure_cpu} 
\begin{tabular}{c|c|c}
\hline
\hline
Virtual Machine (Azure) & A2 standard & D4 standard \\
\hline 
price & $\$ 0.120/hour$ & $\$ 0.616/hour$ \\
OS & Ubuntu 14.04 & Ubuntu 14.04 \\
Memory & 3.5GB & 28GB  \\
CPU &    8 cores @ 2.20GHz & 2 cores @ 2.20GHz \\
Keras version & 1.1.1  & 1.1.1\\
Theano version & 0.9.0dev & 0.9.0dev \\
\hline
\multicolumn{3}{c}{Time (s)} \\
Pulling trajectories from redis & 2.71 & 1.64 \\
Converting trajectories into flows & 9.65 & 6.05 \\
Predicting the crowd flows & 5.79 & 2.97 \\
Pushing results into redis & 0.41 & 0.27 \\
\hline
Total & 18.56 & 10.93 \\
\hline
\hline
\end{tabular}
\end{center}
\end{table}%
\section{Related Work}\label{sec:related_work}
\subsection{Crowd Flow Prediction} 
There are some previously published works on predicting an individual's movement based on their location history \cite{Fan2015,Song2014}. 
They mainly forecast millions, even billions, of individuals' mobility traces rather than the aggregated crowd flows in a region. 
Such a task may require huge computational resources, and it is not always necessary for the application scenario of public safety. 
Some other researchers aim to predict travel speed and traffic volume on the road \cite{Abadi2015IToITS,Silva2015PotNAoS,Xu2014IToITS}. Most of them are predicting single or multiple road segments, rather than citywide ones \cite{Xu2014IToITS,Chen2014}. 
Recently, researchers have started to focus on city-scale traffic flow prediction \cite{Hoang2016,Li2015}. 
Both work are different from ours where the proposed methods naturally focus on the individual region not the city, and they do not partition the city using a grid-based method which needs a more complex method to find irregular regions first. 

\subsection{Classical Models for Time Series Prediction}
Predicting the flows of crowds can be viewed as a type of time series prediction problem. 
There are several conventional linear models for such problem. 
The historical average model is portable, which simply uses the average value of historical time series to predict future value of time series. However, the model unable to respond to dynamic changes, such as incidents \cite{smith1997traffic}. 
The Auto-Regressive Integrated Moving Average (ARIMA) model assumes that the future value of time series is a linear combination of previous values and residuals, furthermore, in order to obtain stationarity, the nonstationary time series should be differenced before analysis \cite{box2015time}. ARIMA is not suite for time series with missing data, since they relying on uninterrupted time series, and data filling technique might be problematic as the complexity of the situation increase \cite{smith2002comparison}. The additional seasonal difference is often applied to seasonal time series to obtain stationarity before ARIMA being used, which is called SARIMA. The disadvantage of SARIMA is time consuming \cite{smith2002comparison}. The Vector Autoregressive (VAR) models capture the linear inter dependencies among interrelated time series \cite{chandra2009predictions}. However, the correlation between predicted values and residuals is neglected. 

Being different from the above linear models, the artificial neural network (ANN) model is a nonlinear model and commonly used in time series prediction \cite{florio1996neural,dougherty1997short,zhang2003time}.
ANNs have excellent nonlinear modeling ability, but not enough for linear modeling ability \cite{zhang2005neural}. 

\subsection{Deep Neural Networks}
Neural networks and deep learning \cite{lecun2015deep,schmidhuber2015deep,bengio2015deep} have gained numerous success in the fields such as compute vision \cite{Krizhevsky2012,ren2015faster}, speech recognition \cite{graves2013speech,yu2012automatic}, and natural language processing \cite{le2014distributed}. 
For example, convolutional neural networks won the ImageNet \cite{ILSVRC15} competition since 2012, and help AlphaGo \cite{silver2016mastering} beat Go human champion\footnote{\url{https://en.wikipedia.org/wiki/AlphaGo_versus_Lee_Sedol}}. 
Recurrent neural networks (RNNs) have been used successfully for sequence learning tasks \cite{sutskever2014sequence}. The incorporation of long short-term memory (LSTM) \cite{hochreiter1997long} or gated recurrent unit (GRU) \cite{cho2014learning} enables RNNs to learn long-term temporal dependency.  However, both kinds of neural networks can only capture spatial or temporal dependencies. Recently, researchers combined above networks and proposed a convolutional LSTM network \cite{xingjian2015convolutional} that learns spatial and temporal dependencies simultaneously. Such a network cannot model very long-range temporal dependencies (\eg{}, period and trend), and training becomes more difficult as depth increases. 

In our previous work \cite{Zhang2016}, a general prediction model based on DNNs was proposed for spatio-temporal data. In this paper, to model a specific spatio-temporal prediction (\ie{} citywide crowd flows) effectively, we mainly propose employing the residual learning and a parametric-matrix-based fusion mechanism. A survey on data fusion methodologies can be found at \cite{Zheng2015Itobd}. 

\subsection{Urban Computing}
Urban computing \cite{Zheng2014AToISaTT}, has emerged as a new research area, which aims to tackle urban challenges (\eg{}, traffic congestion, energy consumption, and pollution) by using the data that has been generated in cities (\eg{}, geographical data, traffic flow, and human mobility). 
A branch of research also partitions a city into grids, and then studies the traffic flow in each region of the city, such as predicting urban air quality \cite{Zheng:2013UAir,Zheng2015}, detecting anomalous traffic patterns \cite{pang2013detection}, inferring missing air quality \cite{XiuwenYi2016}, forecasting of spatio-temporal data \cite{Zhang2016}. 
Besides, some researchers started to research on deep learning methods for urban computing applications.
For example, Song et al. proposed a recurrent-neural-network-based model to predict the person's future movement \cite{song2016deeptransport}. 
Chen et al. proposes a deep learning model to understand how human mobility will affect traffic accident risk \cite{chen2016learning}. 
Both work are very different from ours in terms of approach and problem setting. 
To the best of our knowledge, in the field of urban computing, end-to-end deep learning for forecasting citywide crowd flows has never been done.  

\section{Conclusion and Future Work}
We propose a novel deep-learning-based model for forecasting the flow of crowds in each and every region of a city, based on historical spatio-temporal data, weather and events. 
Our ST-ResNet is capable of learning all spatial (\textit{nearby} and \textit{distant}) and temporal (\textit{closeness}, \textit{period}, and \textit{trend}) dependencies as well as external factors (\eg{} weather, event). 
We evaluate our model on two types of crowd flows in Beijing and NYC, achieving performances which are significantly beyond 9 baseline methods, confirming that our model is better and more applicable to the crowd flow prediction. The code and datasets have been
released at: {https://www.microsoft.com/en-us/research/publication/deep-spatio-temporal-residual-networks-for-citywide-crowd-flows-prediction}.
We develop a Cloud-based system, called UrbanFlow, that can monitor the real-time crowd flows and provide the forecasting crowd flows in near future using our ST-ResNet. 

In the future, we will consider other types of flows (\eg{}, metro card swiping data, taxi/truck/bus trajectory data, and phone signals data), and use all of them to generate more types of flow predictions, and \textit{collectively} predict all of these flows with an appropriate fusion mechanism. 

\section{Acknowledgments}
This work was supported by the National Natural Science Foundation of China (Grant No. 61672399 and No. U1401258), and the China National Basic Research Program (973 Program, No. 2015CB352400). 

\bibliography{refAI}
\end{document}